%% file: main.tex
\documentclass[10pt,twocolumn,letterpaper]{article}

\usepackage[pagenumbers]{wacv} 

\usepackage{graphicx}
\usepackage{amsmath}
\usepackage{amssymb}
\usepackage{booktabs}

\usepackage{multirow}
\usepackage{makecell}
\usepackage{pifont}

\usepackage{algorithm}
\usepackage{algpseudocode}

\usepackage[pagebackref,breaklinks,colorlinks]{hyperref}

\usepackage[capitalize]{cleveref}
\crefname{section}{Sec.}{Secs.}
\Crefname{section}{Section}{Sections}
\Crefname{table}{Table}{Tables}
\crefname{table}{Tab.}{Tabs.}

\def\wacvPaperID{1953} 
\def\confName{WACV}
\def\confYear{2025}

\begin{document}

\title{Data-Efficient 3D Visual Grounding via Order-Aware Referring}

\author{Tung-Yu Wu$^{1}$\thanks{Equal Contribution} \hspace{4.0mm} Sheng-Yu Huang$^{1*}$ \hspace{4.0mm} Yu-Chiang Frank Wang$^{1,2}$
\\ $^{1}$Department of Electrical Engineering, National Taiwan University \\ 
$^{2}$NVIDIA\\
{\tt\small \{b08901133, f08942095\}@ntu.edu.tw, frankwang@nvidia.com}}

\maketitle


\input{0_Abstract}
\input{1_Introduction}
\input{2_Related_Works}
\input{3_Method}
\input{4_Experiments}
\input{5_Conclusions}

{\small
\bibliographystyle{ieee_fullname}
\bibliography{egbib}
}

\clearpage
\section*{Supplementary Material}

\appendix
\input{supp/1_implementation_details}
\input{supp/3_additional_quantitative}
\input{supp/4_loss_ablation}
\input{supp/6_order_length_as_subsets}

\input{supp/5_more_visualization}
\input{supp/2_prompt_and_examples}

\input{supp/7_limitations}

\end{document}

%% file: 0_Abstract.tex
\begin{abstract}
3D visual grounding aims to identify the target object within a 3D point cloud scene referred to by a natural language description. Previous works usually require significant data relating to point color and their descriptions to exploit the corresponding complicated verbo-visual relations. In our work, we introduce \textit{Vigor}, a novel Data-Efficient 3D \textbf{Vi}sual \textbf{G}rounding framework via \textbf{O}rder-aware \textbf{R}eferring. Vigor leverages LLM to produce a desirable referential order from the input description for 3D visual grounding. With the proposed stacked object-referring blocks, the predicted anchor objects in the above order allow one to locate the target object progressively without supervision on the identities of anchor objects or exact relations between anchor/target objects. We also present an order-aware warm-up training strategy, which augments referential orders for pre-training the visual grounding framework, allowing us to better capture the complex verbo-visual relations and benefit the desirable data-efficient learning scheme. Experimental results on the NR3D and ScanRefer datasets demonstrate our superiority in low-resource scenarios. In particular, Vigor surpasses current state-of-the-art frameworks by 9.3\% and 7.6\% grounding accuracy under 1\% data and 10\% data settings on the NR3D dataset, respectively. Our code is publicly available at \url{https://github.com/tony10101105/Vigor}.
\end{abstract}

%% file: 1_Introduction.tex
\section{Introduction}
\label{sec:intro}

Visual grounding is an emerging task that aims to ground a target object in a given 2D/3D scene from a natural description, where the description contains information to identify the target object (e.g., color, shape, or relations to other anchor objects). This task is potentially related to industrial applications to AR/VR and robotics~\cite{lee2021all, kochanski2019image, anderson2018vision}. Compared to object detection, the main challenge of visual grounding lies in the requirement to find \textit{the only one} object described in the given natural description, while there might be multiple objects with the same class of the target object appearing in the scene. Therefore, the model is expected to identify the relations between all objects in the scene to find the ideal target object according to the given description. In recent years, significant progress has been made in image-based 2D visual grounding~\cite{kamath2021mdetr, zhang2022glipv2, liu2023grounding, shtedritski2023does, yang2024fine, zhao2023bubogpt}. However, comparatively fewer efforts are directed towards addressing the more intricate challenge of 3D visual grounding, raised from joint consideration of the unstructuredness of natural language descriptions and scattered object arrangements in the 3D scene. The complications of the two modalities make it challenging to directly refer to the target object with plain cross-modal interaction between the features of the scene point cloud and the referring description, showing the need for additional research in the field of 3D visual grounding.

As pioneers of 3D visual grounding, Referit3D~\cite{achlioptas2020referit3d} and ScanRefer~\cite{chen2020scanrefer} are two benchmark datasets that build upon the point cloud scene provided by the ScanNet~\cite{dai2017scannet} dataset. The former presents a graph neural network (GNN)~\cite{scarselli2008graph}-based framework to explicitly learn the object spatial relations as the baseline. The latter designs a verbo-visual cross-modal feature extraction and fusion pipeline as the baseline. Following the settings of the aforementioned benchmarks, several subsequent methods are presented~\cite{huang2022multi, yang2021sat, bakr2022look, luo20223d, jain2022bottom, feng2021free, huang2021text, chen2022language, wang-etal-2023-3drp}. 
However, these approaches use a referring head to localize the target object directly. Without explicitly considering any additional information about the anchor objects mentioned in the description, models must implicitly discover the relation between the anchor objects and the target object. They may be misled by other similar objects presented, as pointed out in~\cite{bakr2023cot3dref}. To overcome this issue, some approaches~\cite{yuan2022toward, abdelreheem2024scanents3d, wu2023eda} propose to incorporate anchor objects during training by including their label annotations~\cite{hsu2023ns3d, bakr2023cot3dref}. Nonetheless, human annotators are usually required to obtain this additional linguistic information~\cite{yuan2022toward, abdelreheem2024scanents3d}, causing potential difficulty in scaling up to larger datasets for real-world applications.

\input{figures/teaser}

To eliminate the need of human annotation for training grounding models, recent approaches~\cite{zhu2024open3dseg, takmaz2023openmask3d,hsu2023ns3d, bakr2023cot3dref} leverage pre-trained 2D priors (e.g., SAM~\cite{sam}, LDM~\cite{rombach2022ldm}) or large language models (LLMs) for automatic dissection of the descriptions and generation of prior linguistic knowledge. For example, Diff2Scene~\cite{zhu2024open3dseg} conducts the use of LDM to obtain text-conditioned 2D semantic maps as pseudo labels to guide a 3D segmentation model to achieve scene understanding and produce zero-shot 3D visual grounding. Unfortunately, limited by the spatial understanding ability of 2D diffusion models between multiple objects as described in~\cite{ma2024directed}, the ability of Diffi2Scene to achieve effective visual grounding for complex description with multiple anchor objects is still unclear. On the other hand, NS3D~\cite{hsu2023ns3d} utilizes Codex~\cite{chen2021evaluating} to parse descriptions into nested expressions and designs a neuro-symbolic framework to find the target object step-by-step. However, it only considers fixed-template relations between objects (e.g., below/above, near/far, etc.) and cannot be easily extended to arbitrary natural descriptions. Inspired by the mechanism of human perception system~\cite{mcvay2009conducting, chen2017unified}, CoT3DRef~\cite{bakr2023cot3dref} generates the referential order of a description that points from anchor objects to the final target object using LLM. For example, for a description ``\textit{Find the water bottle on the table nearest to the door.}'', the referential order is generated as \{ ``\textit{door}'' (anchor), ``\textit{table}'' (anchor), ``\textit{water bottle}'' (target) \}. Additionally, it utilizes a rule-based algorithm to localize the identities of the above anchor/target objects, which guides a transformer-based module to predict the final target object. However, as noted in~\cite{bakr2023cot3dref}, such rule-based identity prediction might not be applicable for scenarios with complex language descriptions.



In this paper, we propose a data-efficient 3D \textbf{Vi}sual \textbf{G}rounding framework via \textbf{O}rder-aware \textbf{R}eferring (Vigor). Leveraging the LLM-parsed referential order, Vigor exploits the awareness of anchor objects from the textual description, as depicted in Fig.~\ref{fig:teaser}. With such ordered anchor objects as guidance, a series of Object Referring blocks are deployed to process the corresponding objects, each performing feature enhancement to update the visual features of corresponding objects. Since only the ground-truth grounding information of the target object is available during training (no ground-truth referential order observed), we additionally introduce a unique warm-up learning strategy to Vigor. This pre-training scheme can be viewed as augmenting object labels and referential orders to initialize Vigor so that it can be realized in data-efficient training schemes. Our experiments on real-world benchmark datasets confirm that Vigor performs favorably against recent 3D visual grounding methods, especially when the size of training data is limited.

We now summarize our contribution as follows:
\begin{itemize}
\item We present a Data-Efficient 3D \textbf{Vi}sual \textbf{G}rounding Framework via \textbf{O}rder-Aware \textbf{R}eferring (Vigor), which performs 3D visual grounding from natural description inputs.

\item By utilizing sequential yet consecutive Object Referring blocks, Vigor is able to locate anchor/target objects mentioned in the description by considering plausible referential orders established by LLM.
\item We introduce a warm-up strategy that introduces the model with the ability to locate anchor/target object identities by synthesizing training examples of reliable labels and referential orders.
\item Through comprehensive experiments, we show that Vigor achieves satisfactory performances in various low-source settings, surpassing current grounding approaches significantly.
\end{itemize}

%% file: figures/teaser.tex
\begin{figure*}[tb]
  \centering
  \includegraphics[width=0.8\textwidth]{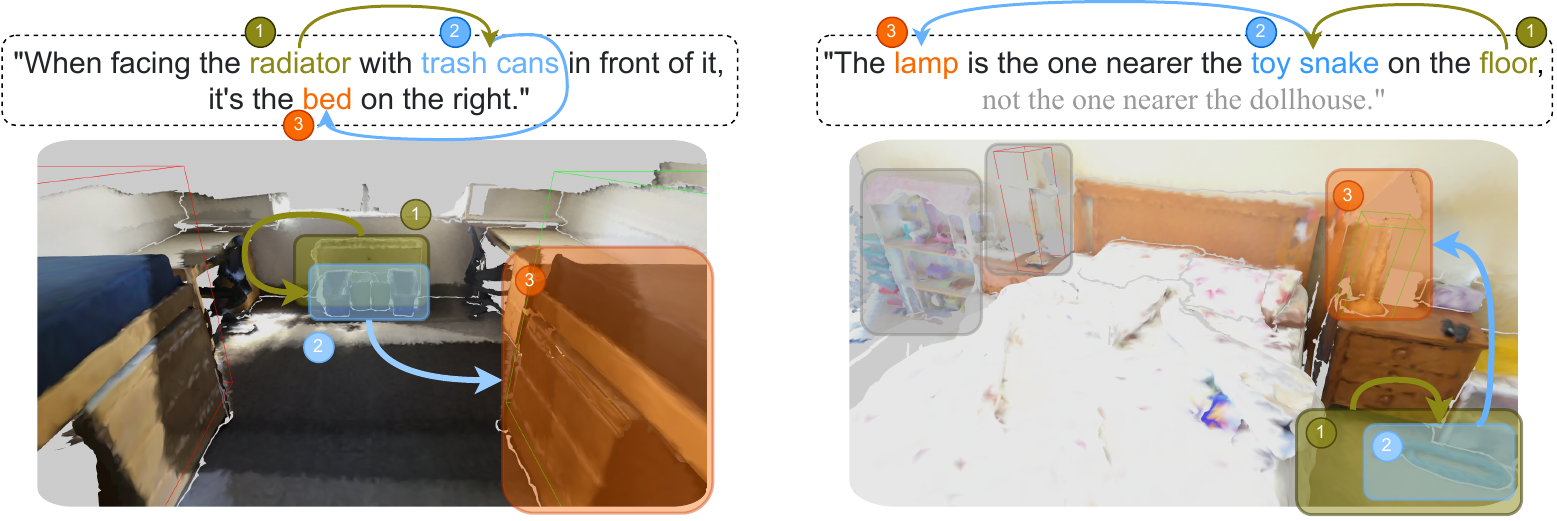}
  \vspace{-0mm}
  \caption{\textbf{Referential orders for 3D grounding.} The order manifests an anchor-to-target referring process that helps the grounding model identify the target object described in the input.}
  \vspace{-2mm}
  \label{fig:teaser}
\end{figure*}


%% file: 2_Related_Works.tex
\section{Related Work}
\subsection{2D Visual Grounding}
2D visual grounding aims to locate the target object in an image referred to by a natural language description, with various approaches being proposed in recent years~\cite{kamath2021mdetr, li2022grounded, zhang2022glipv2, liu2023grounding, yao2021cpt, subramanian-etal-2022-reclip, shtedritski2023does, yang2024fine}. Among them, verbo-visual feature alignment frameworks~\cite{kamath2021mdetr, li2022grounded, zhang2022glipv2, liu2023grounding} have proven themselves to be an effective way to equip models with abilities to tackle description contexts and image semantics simultaneously. Particularly, as one of the pioneers, MDETR~\cite{kamath2021mdetr} extends DETR~\cite{carion2020end}, an end-to-end object detection framework, to incorporate text modalities with the proposed text-image alignment contrastive losses. GLIP~\cite{zhang2022glipv2} takes a step forward to improve the performance of visual grounding by proposing unified multi-task learning that includes object localization and scene understanding tasks, showing that these tasks could gain mutual benefits from each other. Grounding DINO~\cite{liu2023grounding} further designs the large-scale grounding pretraining for DINO~\cite{zhang2022dino}, reaching the capability of open-set grounding. Although great progress is achieved, extending these 2D visual grounding methods to 3D scenarios is not easy due to the additional depth information in 3D data that triggers more complicated object arrangements and more complex descriptions to describe the relations between objects, leaving 3D visual grounding as an unsolved research area.

\subsection{3D Visual Grounding}
In 3D visual grounding, models are designed to jointly handle complicated natural language descriptions and scattered objects within a point cloud scene. Previous approaches attempt to solve this task by either constructing text-point-cloud feature alignment frameworks~\cite{jain2022bottom}, designing pipelines to better exploit the 3D spatial relations of objects~\cite{yuan2021instancerefer, achlioptas2020referit3d, feng2021free, huang2021text, yang2024exploiting, chen2022language}, or bringing in auxiliary visual features~\cite{yang2021sat, bakr2022look, luo20223d, huang2022multi}. Specifically, BUTD-DETR~\cite{jain2022bottom} extends MDETR~\cite{kamath2021mdetr} to 3D visual grounding by adapting a text-point-cloud alignment loss to pull the features of point cloud and text together. To exploit the 3D spatial relations, graph-based methods~\cite{yuan2021instancerefer, achlioptas2020referit3d, feng2021free, huang2021text} utilize GNNs to model the 3D scene, with nodes and edges being the objects and object-to-object relations, to learn their correlations explicitly. Also, some studies craft specialized modules~\cite{he2021transrefer3d, yang2024exploiting, chen2022language, wang-etal-2023-3drp}, such as the spatial self-attention presented in ViL3DRel~\cite{chen2022language} and relation matching network in CORE-3DVG~\cite{yang2024exploiting}, to capture spatial relations among objects. 

To better identify the target object,~\cite{yang2021sat, bakr2022look, luo20223d, huang2022multi} aim to produce richer input semantic information for learning the grounding model. For example, \cite{yang2021sat, bakr2022look, luo20223d} introduce image features by acquiring 2D images of the scene to obtain more color/shape information. MVT~\cite{huang2022multi} projects the point cloud into multiple views for more position information. Although these approaches have achieved great progress in dealing with scattered object arrangements, such methods typically extract a global, sentence-level feature~\cite{devlin-etal-2019-bert, liu2019roberta} from the given natural description. As a result, detailed information such as the target object, anchor objects, and their relations may not be preserved and leveraged properly, potentially reducing training efficiency and prediction accuracy as discussed in~\cite{bakr2023cot3dref}.

To address the above issue, some works have put their efforts into mining the natural descriptions to acquire additional prior knowledge for improved learning~\cite{yuan2022toward, abdelreheem2024scanents3d, wu2023eda, hsu2023ns3d, bakr2023cot3dref}. Specifically, ScanEnts3d~\cite{abdelreheem2024scanents3d} and 3DPAG~\cite{yuan2022toward} recruit human annotators to establish one-to-one matching between each anchor object mentioned in the description and the corresponding object entity in the 3D scene. With such additional information, they design dense word-object alignment losses to improve the training. However, annotating one-to-one text-3D relations requires considerable labor effort. For example, it takes more than 3600 hours of workforce commitment in 3DPAG to annotate anchor objects for 88k descriptions.

\subsection{Data-Efficient 3D Visual Grounding}
To eliminate the need for human annotators for learning grounding models, NS3D~\cite{hsu2023ns3d} makes the first attempt to use the LLM. It leverages Codex~\cite{chen2021evaluating} to parse fixed-template descriptions into nested logical expressions, followed by a neural-symbolic framework to execute the logical expressions implemented as programmatic functions. By doing so, NS3D correctly locates the anchor/target objects mentioned in the given description and achieves impressive performance with only $0.5\%$ training data on synthetic datasets.
\input{figures/main_framework}
Unfortunately, NS3D is not designed to handle arbitrary natural descriptions, resulting in complicated expressions and unforeseen functions and hindering the framework from successful execution~\cite{hsu2023ns3d}. Recently, CoT3DRef~\cite{bakr2023cot3dref} proposes to utilize LLMs to acquire the referential order of the description, listing from anchor objects to the final target object. It deploys a rule-based searching method using a traditional sentence parser~\cite{schuster2015generating} to construct the one-to-one matching between class names in the order and potential anchor/target objects in the scene at once. The matching information is encoded by positional encoding as pseudo labels of anchor/target objects and as additional inputs for the proposed transformer-based CoT module to learn the referring process implicitly and predict the final target object. By the above design, CoT3DRef is able to reduce the training data to $10\%$ while preserving competitive performance on real-world datasets. Nevertheless, since the parsed pseudo labels may not be accurate, taking them as additional inputs might result in noisy information and degrade the grounding process, impeding correct target prediction and affecting training stability.


%% file: figures/main_framework.tex
\begin{figure*}[tb]
  \centering
  \includegraphics[width=\textwidth]{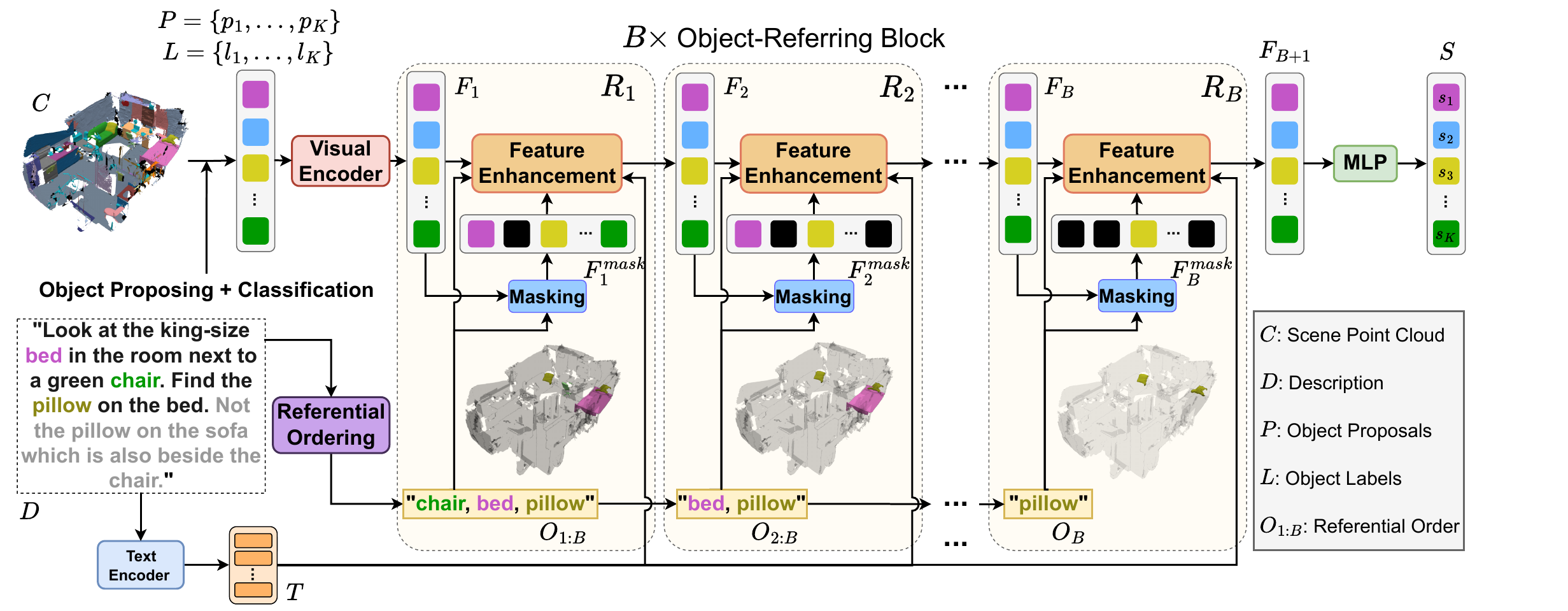}
  \caption{\textbf{Architecture of our 3D Visual Grounding Framework with Order-Aware Referring (Vigor).} By taking a point cloud scene $C$ and a natural description $D$ as inputs, our Vigor produces a referential order of anchor/target objects $O_{1:B}$ and conduct Object-Referring blocks $R_{1:B}$ to locate the target object progressively.}
  \label{fig:main framework}
\end{figure*}

%% file: 3_Method.tex
\section{Methodology}
\label{sec:method}
\subsection{Problem Formulation and Model Overview}
\label{sect:figures}
\subsubsection{Problem formulation}
We first define the setting and notations used in this paper. For each indoor scene, we have a set of colored point cloud $C \in \mathbb{R}^{N \times 6}$, where $N$ denotes the number of the points in the scene, with each point represented in terms of its three-dimensional coordinate and RGB spaces. $C$ is processed to acquire $K$ object proposals $P=\{\boldsymbol{p}_{1}, \ldots, \boldsymbol{p}_{K}\}$
that represent possible objects in the scene, with each proposal containing $I$ points (i.e., $\boldsymbol{p}_{n} \in \mathbb{R}^{I \times 6}$, $n \in [1, \cdots, M]$). $P$ is obtained either by pre-trained object segmentation networks~\cite{jiang2020pointgroup, qi2019deep, wu20223d} or directly from the dataset. Along with $P$, the class labels $L=\{l_1, \cdots, l_K\}$ for all proposals are additionally predicted by a Pointnet++~\cite{qi2017pointnet++} classifier.
For 3D grounding, a text description $D$ is given, illustrating the target object in $C$ by describing its color, shape, or relations to other anchor objects. Given the above inputs, our goal is to identify the exact target object that matches $D$ among all objects in the scene by predicting a $K$-dimensional confidence score $S = \{s_1, \cdots, s_K \}$ for classifying the target object. 

\subsubsection{Model overview}
\label{overall pipeline}
 As shown in Fig.~\ref{fig:main framework}, Vigor is composed of $B$ consecutive Object-Referring blocks $\{R_1, \cdots, R_B\}$ to progressively locate the target object. In particular, by taking both $P$ and $D$ as the inputs, Vigor utilizes the Object-Referring blocks $R_{1:B}$ in Fig.~\ref{fig:main framework} to sequentially produce anchor objects to guide the grounding process. Each $R_i$ 
 takes the object feature $F_i \in \mathbb{R}^{K \times d_i}$ and the text feature $T \in \mathbb{R}^{(|D|+1) \times 768}$ as the inputs. Note that $T$ contains a ${1 \times 768}$-dimensional sentence-level feature and ${|D| \times 768}$-dimensional word-level feature, where $|D|$ denotes the length of $D$ after tokenization. By observing $F_i$ and $T$, the Object-Referring block $R_i$ aims to produce the refined feature $F_{i+1}$ along with the updated anchor/target objects and their relations for grounding purposes.
 
To enable our Object-Referring blocks to capture proper information about the anchor/target objects, we apply a Large Language Model (LLM) to $D$ to generate a Referential Order $O_{1:B} = \{O_1, \cdots, O_B\}$ that mimics human perception system of searching target object~\cite{mcvay2009conducting, chen2017unified} by extracting and arranging the class names of the anchor and target objects, similar to~\cite{bakr2023cot3dref}. Specifically, $\{O_1, \cdots, O_{B-1}\}$ represent the class names of the anchor objects, and $O_B$ is the class name of the target object (please refer to Supp.~\textcolor{red}{F} for details of Referential Order generation). Note that $O_{i:B}$ is observed by the $i$-th Object-Referring block $R_i$ as guidance, which updates the features of anchor/target objects with the proposed Feature Enhancing (FE) module. Since the ground truth referential order is \textit{not} available during training, we introduce a unique warm-up strategy for training Vigor. This is achieved by synthesizing \textit{accurate} referential order and anchor/target object labels. It is worth noting that, with the above design, Vigor can be applied for 3D grounding tasks and achieve satisfactory performances with a respectively limited amount of training data. We now detail the design of our Vigor in the following subsections.

\input{figures/order_generation_and_transformer_block}

\subsection{3D Visual Grounding with Order-Aware Object Referring}
\label{order-aware referring}
\subsubsection{Object-referring blocks.}
Given the object proposals $P$, the corresponding labels $L=\{l_1, \cdots, l_K\}$, the encoded text features $T$, and the derived referential order $O_{1:B}$ the as inputs, our Vigor deploys a series of Object-Referring blocks $\{R_1, \cdots, R_B\}$ to perform the grounding task. As depicted in Fig.~\ref{fig:main framework}, this referring process is conducted by leaving out an anchor object and updating the visual features in each step until only the final step locates the target object of interest. 
Thus, the deployment of Object-Referring blocks allows one to focus on the anchor/target objects so that their visual features and spatial relations between them can be exploited while those of irrelevant objects are disregarded. 

Take the $i$-th referring block observing $O_{i:B}$ for example, a \textit{masked feature} $F^{mask}_i = F_i \odot M_i$ is derived by applying a Hadamard product between $F_i$ and a $K$-dimensional binary mask $M_i$ to replace features of object proposals in $F_i$ not belonging to any of the object classes in $O_{i:B}$. 
Such a masking strategy ensures that $F^{mask}_i$ contains objects described in $O_{i:B}$ and hence explicitly suppresses the effects of irrelevant objects that are not in our interests. Thus, the $j$-th entry of $M_i$  (denoted as $m_{ij}$) is defined as:
\begin{equation}
  m_{ij}=
  \begin{cases}
      1 & \text{if class name of } l_j \text{ is in } O_{i:B} ,\\
      0 & \text{otherwise}.
  \end{cases}
  \label{eq:mask_definition}
\end{equation}

\noindent $F_i$ and $F^{mask}_i$ are refined into $F_{i+1}$ for the next referring block via the feature enhancement module (as discussed later). At the final stage, the output $F_{B+1}$ of $R_B$ is utilized to predict the confidence score $S$ that represents the identity of the target object, supervised by a cross-entropy loss $\mathcal{L}_{ref}$. 
Additionally, to ensure the text feature $T$ properly describing the anchor/target objects, we follow CoT3DRef~\cite{bakr2023cot3dref} and apply the language classification loss $\mathcal{L}_{text}$ to $T$ for matching the associated class labels.

\subsubsection{Object feature enhancement}
With the above masking process, each object-referring block is expected to update the object features related to the anchor and target objects. This is realized by our attention-based Feature Enhancement (FE) module. To be more precise, in order to update the features associated with the anchor and target objects in $R_i$ according to $F_i$, $F^{mask}_i$, $T$ and $O_{i:B}$, our FE module aims to exploit their visual features and spatial relations through attention mechanisms. 

Take the FE module in $R_1$ as an example, as depicted in Fig.~\ref{fig:transformer block}, we start with the lower branch which locally emphasizes the features of the potential anchor/target objects via a cross-attention layer by treating $F^{mask}_i$ as value/key and encoded text feature of $O_{1:B}$ (denoted as $T_{O_{1:B}}$) as query. On the other hand, the upper branch of Fig.~\ref{fig:transformer block} explores the spatial relations between all objects by treating the self-attended $F_1$ as the key/value of another cross-attention, with the concatenation of $T_{O_{1:B}}$ and $T$ being query. Finally, an additional cross-attention layer is applied to the previous output features of both branches to obtain the enhanced proposal feature $F_2$, which enriches not only the information of anchor/target objects but also the relations between them. 

On the other hand, to prevent the information extracted from the anchor/target objects from vanishing (i.e., $F_2$ becomes identical to $F_1$) during FE, we introduce an additional masking loss $\mathcal{L}_{mask}$ by projecting $F_2$ from $K \times d_2$-dimensional to $K \times 1$-dimensional digits with MLPs to classify if each proposal in $F_2$ is previously masked in $F^{mask}_1$. The masking loss $\mathcal{L}_{mask}$ is defined as:
\begin{equation}
  \mathcal{L}_{mask}=\mathcal{L}_{BCE}(MLP(F_{2}), M_1),
  \label{eq:loss mask}
\end{equation}
where $M_1$ represents the $K$-dimensional binary mask as defined in Eqn.~\ref{eq:mask_definition}. It is worth noting that, $\mathcal{L}_{mask}$ is applied to output features of each referring block with a similar formulation to ensure each output feature contains the information of the current anchor/target objects correspondingly. 



\subsection{Order-Aware Warm-up with Synthetic Referential Order}
\label{pretraining}
Although Vigor is designed to produce a referential order of anchor objects for localizing the target object, only the ground truth point cloud information of the target object is given during training. Thus, the above framework is viewed as a weakly-supervised learning scheme since there is \textit{no} ground truth referential order available during training. To provide better training supervision, we warm-up Vigor with a simple yet proper synthetic 3D visual grounding task, where the ground-truth labels of anchor/target objects and descriptions with accurate referential orders can be obtained. This warm-up strategy is presented below.

\subsubsection{Augmenting plausible referential order and description}

To provide better training supervision and to ensure the reliability of our synthesized data, the constructed description and the corresponding referential order need to be easily and uniquely determined based on anchor/target objects. In our work, we choose to consider spatial relations between objects that are independent of viewpoint (e.g., \textit{``nearest''} or \textit{``farthest''}) as the constructing descriptions, suggesting the referential order of anchor/target objects during this data augmentation stage. As highlighted in Fig.~\ref{fig:warmup_pipeline} and Algorithm \textcolor{red}{A1} of our supplementary material, given $P$, $L$, and $B$, we construct an augmented referential order $O^{aug}_{1:B}$ by choosing $B$ different class labels $\{l^{aug}_1, \cdots, l^{aug}_B\}$ from $L$ and extracting their class names. The augmented description $D^{aug}$ is then derived as:


\noindent \textit{``There is a $\{O^{aug}_{1}\}$ in the room, find the $\{O^{aug}_{2}\}$ farthest to it, and then find the $\{O^{aug}_{3}\}$ farthest to that $\{O^{aug}_{2}\}$, \{\ldots\}, finally you can see the $\{O^{aug}_{B}\}$ farthest to that $\{O^{aug}_{B-1}\}$.''}

\noindent Since $D^{aug}$ is constructed following the appearing sequence of object names in $O^{aug}_{1:B}$, it is guaranteed that $O^{aug}_{1:B}$ is a correct referential order w.r.t. $D^{aug}$ and thus can be served as ground truth supervision for pre-training Vigor.


It is worth noting that, to have each object in $D^{aug}$ uniquely defined, we only keep one proposal $p^{aug}_1$ with the class name of $\{O^{aug}_{1}\}$ in $P$ and remove all the other proposals with that class name. As a result, all the anchor and target objects in $P$ (denoted as $p^{aug}_{1:B} = \{p^{aug}_1, \cdots, p^{aug}_B\}$) according to $D^{aug}$ and their corresponding identities are uniquely determined (i.e., $p^{aug}_2$ is assigned by finding the farthest proposal against $p^{aug}_1$ with label $l^{aug}_2$, and the rest of the anchor/target objects are determined consecutively with the same strategy).

\subsubsection{Warm-up objectives}
To have Vigor follow $O^{aug}_{1:B}$ to refer $p^{aug}_i$ in the $i$-th referring block $R_i$, we design a coordinate loss $\mathcal{L}_{crd}$ to encourage the output feature $F_{i+1}$ of $R_i$ to identify the coordinate of all proposals in $P$ w.r.t. $p^{aug}_i$. Thus, we calculate $\mathcal{L}_{crd}$ as:
\begin{equation}
  \mathcal{L}_{crd}=\frac{1}{B}\sum_{i=1}^{B}\mathcal{L}_{MSE}(MLP(F_{i+1}), V-\mathbb{I}\cdot{{v}_{i}}),
  \label{eq:loss coor}
\end{equation}
where $MLP(\cdot)$ represents MLP layers applied to $F_{i+1}$, $V$ is a $K\times3$ matrix representing center coordinates of calculated bounding-boxes of all $K$ proposals in $P$, $\mathbb{I}$ stands for a $K\times1$-dimensional identity vector, and ${v}_{i}$ is the $1\times3$-dimensional center coordinate of $p^{aug}_i$.

We note that, the referential loss $\mathcal{L}_{ref}$ mentioned in Sec.~\ref{order-aware referring} is also extended to classify both the identity of anchor objects using $F_{2:B}$ and the identity of the target object for $F_{B+1}$ during the warm-up process. To this end, we can define the objectives used during our warm-up process by summing up the referential loss $\mathcal{L}_{ref}$ (for both anchor and target objects), the masking loss $\mathcal{L}_{mask}$, the language classification loss $\mathcal{L}_{text}$ and the coordinate loss $\mathcal{L}_{crd}$. With this warm-up strategy, Vigor is initialized to observe relations between anchor/target objects before the subsequent fully-supervised training stage. Later we will verify that, with this pre-training scheme, our Vigor produces satisfactory grounding performances especially when the amount of supervised training data is limited.

\subsection{Overall Training Pipeline}
We now summarize the training of Vigor. With the warm-up stage noted in Sec.~\ref{pretraining}, we take point cloud data with real-world natural descriptions to continue the training process. Since the identity of anchor objects is unknown, we only apply $\mathcal{L}_{ref}$ (for the target object only),  $\mathcal{L}_{mask}$, and $\mathcal{L}_{text}$ as supervision. The overall training pipeline is summarized in Algorithm \textcolor{red}{A2} in supplementary. 

%% file: figures/order_generation_and_transformer_block.tex
\begin{figure*}[t]
  \centering
  \begin{subfigure}{0.35\linewidth}
  \centering
    \includegraphics[width=\textwidth]
    {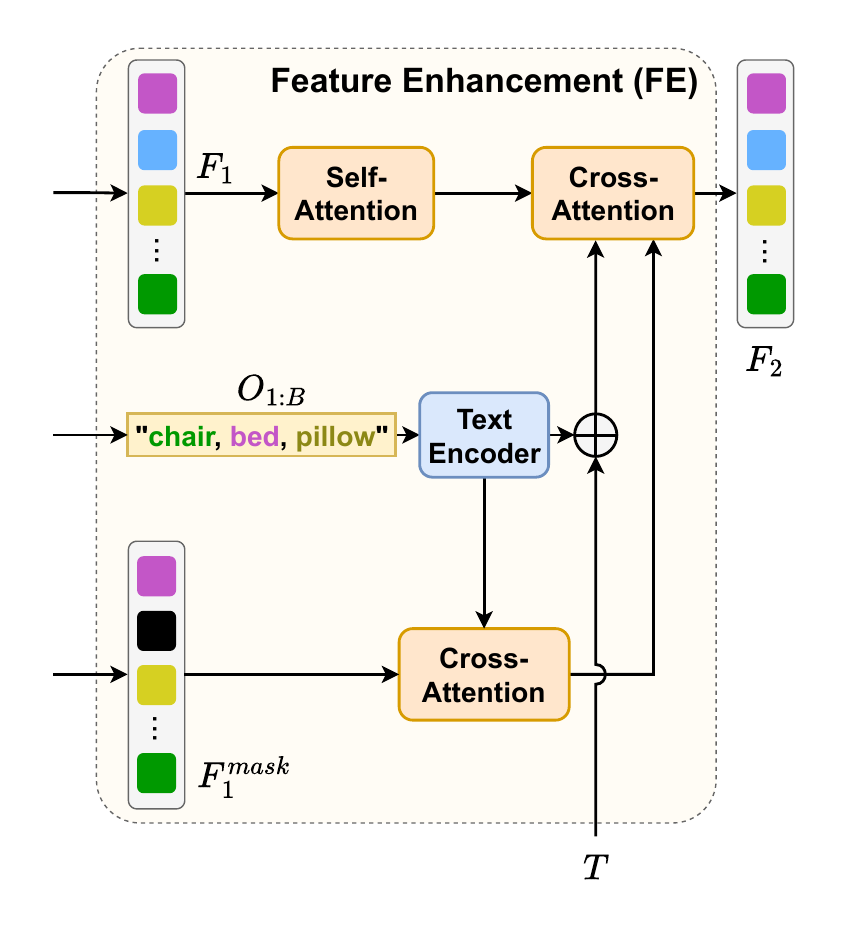}
    \caption{\textbf{Feature Enhancement (FE).} Taking $R_1$ as an example, FE processes the objects described mentioned in $O_{1:B}$ and the relations between them by attending the masked feature $F^{mask}_1$. Thus, only object features related to $O_{1:B}$ would be refined as $F_2$.} 
    \label{fig:transformer block}
  \end{subfigure}
  \hfill
  \begin{subfigure}{0.59\linewidth}
    \includegraphics[width=\textwidth]{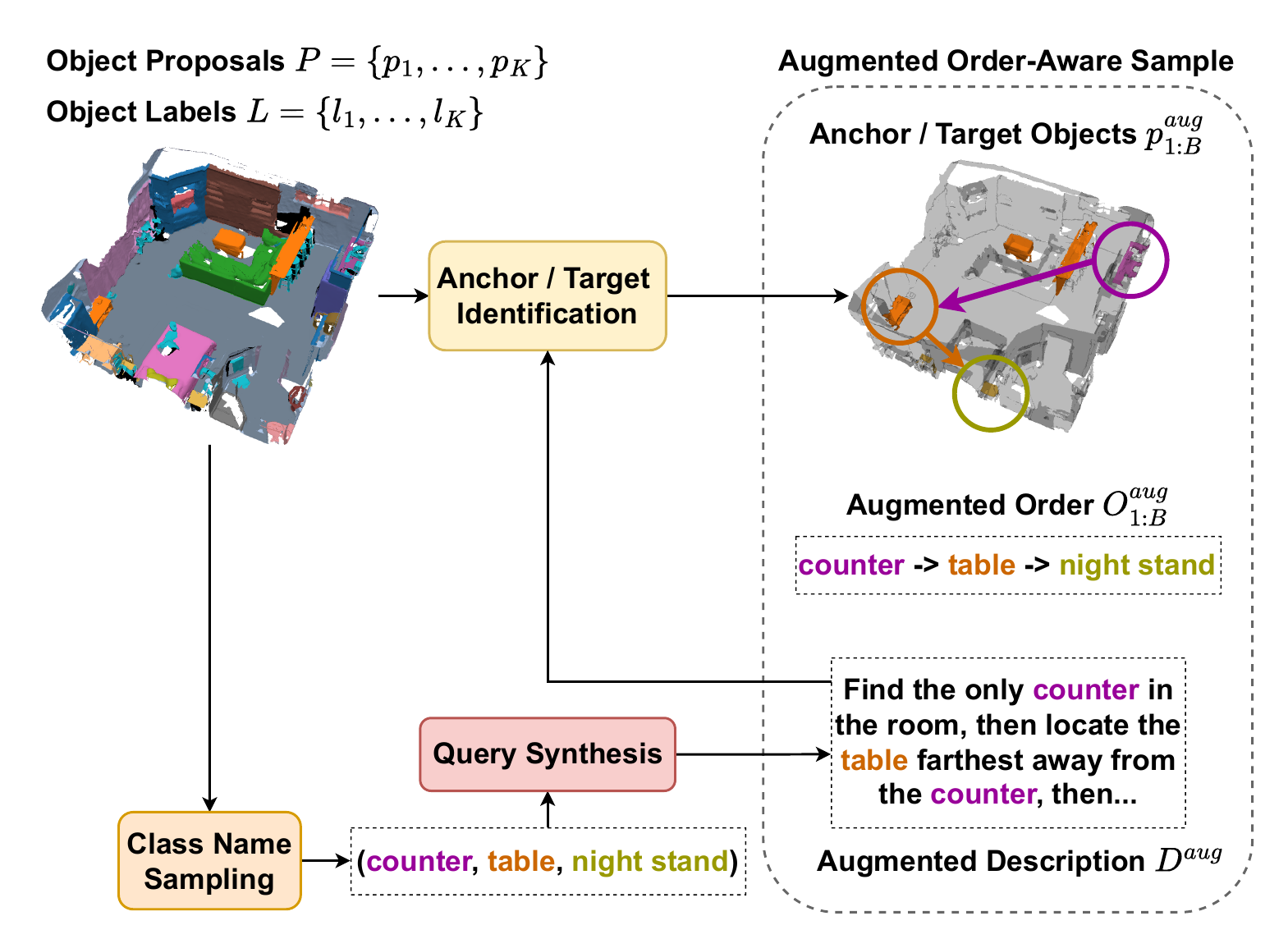}
    \caption{\textbf{Synthesizing a referential order and description for order-aware learning warm-up.}
    Given $P$ and $L$, several class names are sampled to construct $D^{aug}$. The identities of anchor/target objects $p^{aug}_{1:B}$ described in $D^{aug}$ are then located by considering the center coordinates and class name of each proposal. By the above design, the augmented referential order $O^{aug}_{1:B}$ is uniquely determined (i.e., the appearance order of each sampled class name in $D^{aug}$).}

    \label{fig:warmup_pipeline}
  \end{subfigure}
  \caption{Illustration of feature enhancement and synthesizing warmup data in Vigor.}
  \label{fig:block and warmup}
\end{figure*}

%% file: 4_Experiments.tex
\section{Experiments}
\label{experiments}

\input{tables/nr3d_main_table}

\subsection{Dataset}
\paragraph{NR3D}
NR3D~\cite{achlioptas2020referit3d} dataset consists of 707 indoor scenes in ScanNet~\cite{dai2017scannet} with 28715/7485 description-target pairs in the training/testing set, where the descriptions are collected from human annotators. There are 524 different object classes in the scenes in total. NR3D provides ground-truth class-agnostic object proposals, where each point in the scene is properly assigned to its corresponding proposal. As a result, models are only required to classify the target object that uniquely matches the description among all proposals in the scene, with classification accuracy (Acc in \%) being the metric. 

\paragraph{ScanRefer}
ScanRefer~\cite{chen2020scanrefer} contains 36665/9508 description-target pairs across a total of 800 indoor scenes in its training/validation set, where the descriptions are also collected from human annotators. Also derived from ScanNet~\cite{dai2017scannet} but different from NR3D, perfect object proposals are not available in ScanRefer, and therefore, additional object proposers are required for all methods. Nevertheless, Acc in \% under 0.25 and 0.5 intersection over union (IoU) are used as the metrics for ScanRefer. In ScanRefer, since ground-truth object proposals are unavailable, we adopt the visual encoder of M3DRef-CLIP~\cite{zhang2023multi3drefer} that applies PointGroup~\cite{jiang2020pointgroup} to perform object segmentation as proposals $P$ and object classification as labels $L$. We do not perform object-aware pre-training for ScanRefer since imperfect object proposals may lead to noisy synthetic samples and hinder Vigor's training stability. 
\input{tables/scanrefer_main_table}
\subsection{Quantitative Results for Data Efficiency}
We present the quantitative results on NR3D and ScanRefer, with consideration of different amounts of available training samples against several baselines by reproducing from their official implementation. Results of NR3D are shown in Table~\ref{tab:nr3d main table}. Vigor possesses a considerably superior performance when training with 1\% (287), 2.5\% (717), 5\% (1435), and 10\% (2871) NR3D training set samples. Specifically, using only 1\% data, Vigor achieves 33.5 overall Acc, which even surpasses SOTA methods with 10\% of data. This suggests that Vigor is preferable for 3D grounding, especially when paired training data is limited. Table~\ref{tab:scanrefer main table} shows the results on ScanRefer datasets with 5\% (1833) and 10\% (3666) training samples. Vigor still surpasses all baselines under the conditions where imperfect object proposals are used, indicating the robustness of our methods. The detailed performance on different official subsets of NR3D are in the Supp.~\textcolor{red}{B.1}.

\input{tables/component_ablations}
\subsection{Ablation Studies}
\input{figures/qualitative_results}
We ablate different components of our Vigor in Table~\ref{tab:component ablations}, with performances on NR3D under \textbf{1\%}, \textbf{10\%}, and \textbf{100\%} training samples available to explore the effectiveness of each component with different amounts of data. Baseline model A extracts class names of anchor/target objects from $D$ using a language parser~\cite{nltk} and constructs the referential order according to the appearance of the names in $D$ directly. Also, we employ each of our Referring Blocks $R_i$ in model A without the FE module, i.e., directly using $F_i$ to calculate attention with text features of $O_{i:B}$ and $D$. Model B enhances the accuracy on 1\% and 10\% training data by conducting our two-stage object ordering with LLM. When further applying the order-aware pre-training in model C, significant improvements in all settings, especially for 1\% available data, are observed.
By conducting our pre-training strategy, our Vigor is able to learn foundational concepts of ordering and relations to locate the target objects progressively. Finally, our full model in the last row, incorporating FE modules each $R_i$ to enhance features of anchor/target objects with $F^{mask}_i$, achieves optimal results on both settings. This verifies the success of our proposed modules and warm-up strategy, especially when the available training data is very limited.


\subsection{Qualitative Results}
Fig.~\ref{fig:qualitative results} demonstrates the qualitative results of Vigor on NR3D, with MVT being the baseline. We display four successful cases and two failed cases. It is shown that Vigor can successfully identify the target object referred by one to multiple anchor objects, even in lengthy descriptions. Failed cases include those that refer by shapes or have a very small target object that is hard for Pointnet++ to capture visual information.


%% file: tables/nr3d_main_table.tex
\begin{table}[tb]
  \caption{\textbf{Data Efficient Grounding accuracy (\%) on NR3D.} Note that each column shows the results trained with a specific amount of training data.}
  \vspace{-0mm}
  \label{tab:nr3d main table}
  \centering
  \addtolength{\tabcolsep}{8pt}
  \resizebox{\linewidth}{!}{
  \begin{tabular}{@{}l|cccc@{}}
    \toprule
    \multirow{2}{*}{Method} & \multicolumn{4}{c}{Labeled Training Data} \\
    \cline{2-5}
    &1\% & 2.5\% & 5\% & 10\% \\
    \midrule
    Referit3D~\cite{achlioptas2020referit3d} & 4.4 & 13.6 & 20.3 & 23.3 \\
    TransRefer3D~\cite{he2021transrefer3d} & 11.0 & 16.1 & 21.9 & 25.7  \\
    SAT~\cite{yang2021sat} & 11.6 & 16.0 & 21.4 & 25.0 \\
    BUTD-DETR~\cite{jain2022bottom} & \underline{24.2} & \underline{28.6} & 31.2 & 33.3 \\
    MVT~\cite{huang2022multi} & 9.9 & 16.1 & 21.6 & 26.5  \\
    MVT + CoT3DRef~\cite{bakr2023cot3dref} & 9.4 & 17.3 & 26.5 & 38.2  \\
    ViL3DRel + CoT3DRef~\cite{bakr2023cot3dref} & 22.4 & 27.3 & \underline{33.8} & \underline{38.4}  \\
    Vigor (Ours) & \textbf{33.5} & \textbf{36.1} & \textbf{41.5} & \textbf{46.0}  \\
  \bottomrule
  \end{tabular}}
\end{table}

%% file: tables/scanrefer_main_table.tex
\begin{table}[tb]
  \caption{\textbf{Grounding accuracy (\%) on the ScanRefer validation set}. In this table, different amounts of training data are considered.}
  \label{tab:scanrefer main table}
  \centering
  \addtolength{\tabcolsep}{2pt}
  \resizebox{\linewidth}{!}{
  \begin{tabular}{@{}l|cc|cc@{}}
    \toprule
    \multirow{2}{*}{Method} & \multicolumn{4}{c}{Labeled Training Data} \\
    \cline{2-5}
     & \multicolumn{2}{c|}{5\%} & \multicolumn{2}{c|}{10\%} \\
    \cline{2-5}
    & Acc@0.25 & Acc@0.5 & Acc@0.25 & Acc@0.5\\
    \midrule
    ScanRefer~\cite{chen2020scanrefer} & 23.0 & 12.0 & 27.4 & 15.0 \\
    3DVG-Trans.~\cite{zhao20213dvg} & 35.3 & 23.3 & 39.0 & 29.0 \\
    3D-SPS~\cite{luo20223d} & 28.4 & 16.9 & 32.9 & 22.9 \\
    BUTD-DETR~\cite{jain2022bottom} & \underline{38.2} & 26.2 & \underline{40.3} & 28.3 \\
    M3DRef-CLIP~\cite{zhang2023multi3drefer} & 37.2 & \underline{29.9} & 40.0 & \underline{32.4} \\
    Vigor (Ours) & \textbf{39.8} & \textbf{31.5} & \textbf{43.6} & \textbf{34.9} \\
  \bottomrule
  \end{tabular}}
\end{table}

%% file: tables/component_ablations.tex
\begin{table*}[tb]
  \caption{\textbf{Ablation studies of proposed components in Vigor.}
  For methods without LLM Object Ordering, we form $O_{1:B}$ according to the appearance of anchor/target object names in $D$. Cases of 1\%, 10\%, and 100\% training data are considered.
  }
  \label{tab:component ablations}
  \centering
  \addtolength{\tabcolsep}{8pt}
  \resizebox{0.6\textwidth}{!}{
  \begin{tabular}{@{}cccc|ccc@{}}
    \toprule
     & \makecell[c]{FE \\ Module} & \makecell[c]{Order-Aware \\ Pre-training} & \makecell[c]{LLM Object \\ Ordering} & 1\% & 10\% &100\% \\
    \midrule
     A & & & & 8.9 & 35.2&53.9 \\
     B & & & \ding{51} & 10.3&38.8 & 53.8 \\
     C & & \ding{51} & \ding{51} & 25.8&42.5 & 58.0 \\
     \midrule
     Ours& \ding{51} & \ding{51} & \ding{51} & 33.5&46.0 & 59.7 \\
  \bottomrule
  \end{tabular}}
\end{table*}

%% file: figures/qualitative_results.tex
\begin{figure*}[t!]
  \centering
  \includegraphics[width=0.9\textwidth]{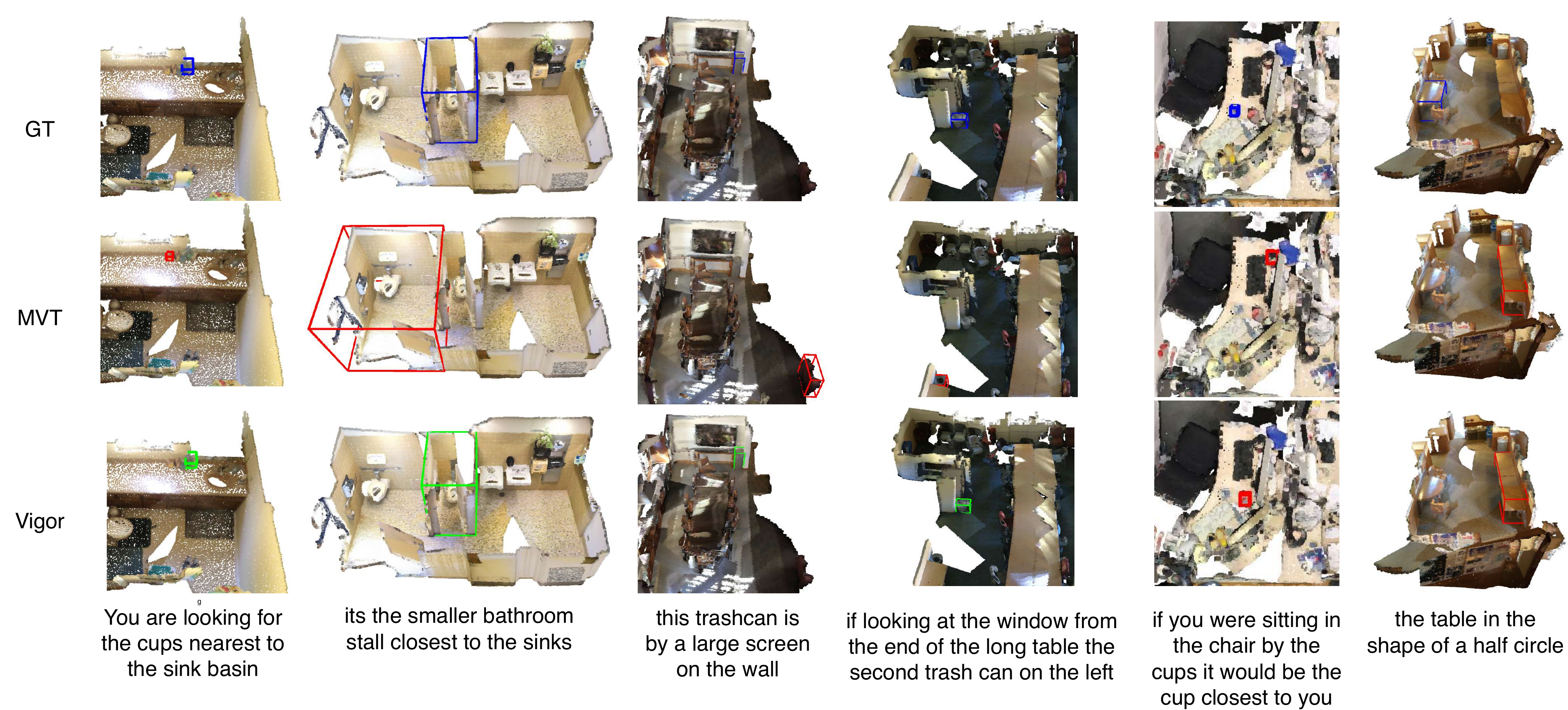}
  \caption{\textbf{3D grounding examples of NR3D.} Note that blue/green/red boxes denote ground truth/correct/incorrect predictions. While both MVT and Vigor fail on the last two cases, it is due to the fact that the size of the target object is extremely small (e.g., cup) and the description does \textit{not} describe any anchor objects.}
  \vspace{-2mm}
  \label{fig:qualitative results}
\end{figure*}

%% file: 5_Conclusions.tex
\section{Conclusions}
We presented a data-efficient 3D \textbf{Vi}sual \textbf{G}rounding Framework with \textbf{O}rder-Aware \textbf{R}eferring (Vigor) in this paper. Vigor identifies anchor/target objects from the LLM-parsed referential order of input description and guides the updates of the associated object features for grounding purposes. The above process is realized by stacked Object-Referring blocks in Vigor, which progressively process the features of the objects of interest in the above referential order. In addition, a unique warm-up scheme to pre-train Vigor was presented, that augments a pseudo yet desirable series of anchor/target objects and enables Vigor to realize relations between objects before formal training. Experiments on benchmark datasets demonstrate that our Vigor performed favorably against SOTA 3D grounding works in a data-efficient manner.

\paragraph{Acknowledgment.}
This work is supported in part by the National Science and Technology Council via grant NSTC 112-2634-F-002-007 and NSTC 113-2640-E-002-003, and the Center of Data Intelligence: Technologies, Applications, and Systems, National Taiwan University (grant nos.113L900902, from the Featured Areas Research Center Program within the framework of the Higher Education Sprout Project by the Ministry of Education (MOE) of Taiwan). We also thank the National Center for High-performance Computing (NCHC) for providing computational and storage resources.

%% file: supp/1_implementation_details.tex
\section{More Details of Vigor}
\label{sup:implementation details}
This section provides more details regarding model implementations and experimental setups. In particular, we first demonstrate the pseudocode of order-aware sample synthesis for pre-training in Sec.~\textcolor{red}{3.4} and the complete training pipeline in Sec.~\textcolor{red}{3.5}. Then, we elaborate on the hyperparameters of Vigor and implementations of baselines for experiments in Sec.~\textcolor{red}{4}.

\subsection{Pre-Training Sample Synthesis and Training Pipeline}
\label{sup:algorithm}
This section provides the pseudocode of synthesizing order-aware samples for Vigor's pre-training in Sec.~\textcolor{red}{3.4} and the complete training pipeline in Sec.~\textcolor{red}{3.5} in our main paper, respectively. Specifically, Algorithm~\ref{alg:data synthesis algorithm} demonstrates the pipeline to synthesize an order-aware pre-training sample given object proposals $P$ and predicted object labels $L$. On the other hand, Algorithm \ref{alg:training algorithm} illustrates the complete pipeline to train Vigor with synthesized samples and natural-description samples, such as NR3D and ScanRefer.
\input{supp/algorithms/pretrain_data_synthesis}

\subsection{Implementation and Details and Experimental Settings}
For both datasets, we use PyTorch~\cite{paszke2019pytorch} library to implement Vigor. We train Vigor using Adam~\cite{adam} optimizer with a single NVIDIA Tesla V100 GPU.

To conduct batch-wise training with a fixed number of Object-Referring blocks, we set the length of referential order $B$ to be 4, i.e., we trim the original referential order from the front if its length exceeds 4 and pad it if its length is lower than 4. We adopt BERT~\cite{devlin-etal-2019-bert} as the text encoder to extract $T$ and Pointnet++~\cite{qi2017pointnet++} as the visual encoder to acquire $F_{1}$. We sample $I$=1024 points for each object proposal in the scene. Object proposal number $K$ and token number $|D|$ are sample-dependent. Object feature dimension $d_{i}$ for $i$-th Object-Referring block is set to 768, aligning with BERT's 768 dimension, to conduct cross-attention.
\input{supp/algorithms/DOrA_training}

\subsubsection{NR3D}
\label{imple details on NR3D}
For NR3D, we use pre-trained Pointnet++ to classify all object proposals as object labels $L$ following BUTD-DETR~\cite{jain2022bottom}. We warm-up Vigor for 15k steps on ScanNet scene point cloud and our augmented samples in Sec.~\textcolor{red}{3.4} and continue on real-world data pairs in NR3D (around 1.2k, 12k, and 120k steps for 1\%, 10\%, and 100\% data, respectively) using a batch size of 24. With one NVIDIA Tesla V100 GPU, the warm-up takes around 3 hours and the NR3D real-world data training takes around 24 hours when training on 100\% data.

\subsubsection{SR3D}
\label{imple details on SR3D}

We present SR3D's results in App.~\ref{sup: sr3d results}. SR3D contains 65844 training samples and 17726 testing samples. Each sample's description is synthesized by simple spatial relations, such as \textit{farthest} and \textit{beside} with simple sentence structures like ``\textit{the monitor that is farthest from the printer.}'' We conduct the object classification and warm-up process as in NR3D and continue on SR3D data pairs, with around 2.7k and 27k training steps for 1\% and 100\% data using batch size 24. SR3D 100\% data training takes around 48 hours with one NVIDIA Tesla V100 GPU.

\subsubsection{ScanRefer}
\label{imple details on scanrefer}
For ScanRefer, following M3DRef-CLIP~\cite{zhang2023multi3drefer}, we use PointGroup~\cite{jiang2020pointgroup} to classify object proposals. We do not apply the warm-up due to noisy and imperfect object proposals and labels under ScanRefer's setting. The batch size is 32, and the GPU is a single NVIDIA Tesla V100 GPU. Training on 100\% data requires around 60 hours.

\subsubsection{Baselines}
For baselines~\cite{achlioptas2020referit3d, he2021transrefer3d, yang2021sat, jain2022bottom, huang2022multi, bakr2023cot3dref, chen2020scanrefer, zhao20213dvg, luo20223d, zhang2023multi3drefer, yang2024exploiting} on NR3D and ScanRefer in Table~\textcolor{red}{1}, ~\textcolor{red}{2}, and ~\textcolor{red}{3}, we utilize their official public implementations with different amounts of available training samples and the full testing set to evaluate their low-resource performance. For the full-data ($100$\%) scenario, we acquire their performance either on the official leaderboard of NR3D/ScanRefer or their published papers.

%% file: supp/algorithms/pretrain_data_synthesis.tex
\begin{algorithm}
	\caption{Order-Aware Sample synthesis for Vigor Pre-training}
	\hspace*{\algorithmicindent}\textbf{Input: }$P$ and $L$\\
	\hspace*{\algorithmicindent}\textbf{Hyperparameters: }$B$\\
    \hspace*{\algorithmicindent}\textbf{Output: }$D^{aug}$, $O^{aug}_{1:B}$, and $p^{aug}_{1:B}$ \\
	\begin{algorithmic}[1]
	    \State randomly sample and arrange $\{l^{aug}_1, \cdots, l^{aug}_B\}$ from $L$.
            \State extract class names of $\{l^{aug}_1, \cdots, l^{aug}_B\}$ as $O^{aug}_{1:B} = \{O^{aug}_{1}, \ldots, O^{aug}_{B}\}$.
            
            \State $D^{aug}=$ \emph{``There is a $\{O^{aug}_{1}\}$ in the room, find the $\{O^{aug}_{2}\}$ farthest to it, and then find the $\{O^{aug}_{3}\}$ farthest to that $\{O^{aug}_{2}\}$, \{\ldots\}, finally you can see the $\{O^{aug}_{B}\}$ farthest to that $\{O^{aug}_{B-1}\}$.''}
            \State get $p^{aug}_{1}$ by randomly removing objects in $P$ with class of $O^{aug}_{1}$ until only one of them is left.
            \State initialize the anchor/target objects set as $\{p^{aug}_1\}$.
		\For {$i= 2,3, \ldots, B$}
                \State for all objects in $P$ with the class of $O^{aug}_{i}$, find the one farthest from $p^{aug}_{i-1}$         \Statex\quad\hspace{0.08em} to be $p^{aug}_{i}$.
                \State append $p^{aug}_{i}$ to $\{p^{aug}_{1}, \ldots, p^{aug}_{i-1}\}$.
		\EndFor
        \State $p^{aug}_{1:B} \leftarrow \{p^{aug}_1, \cdots, p^{aug}_B\}$ 
	\State \textbf{return} $D^{aug}$, $O^{aug}_{1:B}$, and $p^{aug}_{1:B}$
	\end{algorithmic}
\label{alg:data synthesis algorithm}
\end{algorithm}

%% file: supp/algorithms/DOrA_training.tex
\begin{algorithm}
	\caption{Vigor Training Pipeline}
	\hspace*{\algorithmicindent}\textbf{Input: }scene-description paired training samples $\{\{C_1, D_1\}, \{C_2, D_2\}, \ldots\}$\\
	\hspace*{\algorithmicindent}\textbf{Hyperparameters: }pre-training step $S_{p}$, official training step $S_{o}$, and $B$\\
	\begin{algorithmic}[1]
            \State initialize Vigor's model weights $\phi$.
            \State $L_{pre}\leftarrow\{\mathcal{L}_{text}, \mathcal{L}_{mask}, \mathcal{L}_{ref}, \mathcal{L}_{crd}\}$
            \For{$i= 2,3, \ldots, S_{p}$}
    		\State sample a scene point cloud $C$ from paired training samples.
                \State acquire $P$ and $L$ of $C$.
                \State use \ref{alg:data synthesis algorithm} with $\{P, L, B\}$ as inputs to synthesize $\{D^{aug}, O^{aug}_{1:B}, p^{aug}_{1:B}\}$.
                \State use $\{P, L, D^{aug}, O^{aug}_{1:B}, p^{aug}_{1:B}\}$ to update $\phi$ with $L_{pre}$.
            \EndFor
            \State $L_{train}\leftarrow\{\mathcal{L}_{text}, \mathcal{L}_{mask}, \mathcal{L}_{ref}\}$
		\For {$i= 2,3, \ldots, S_{o}$}
    		\State sample a data point $\{C, D\}$ in the training samples
                  \State acquire $P$ and $L$ of $C$.
                  \State apply LLM to acquire $O_{1:B}$.
                  \State use $\{P, L, D, O_{1:B}\}$ to update $\phi$ with $L_{train}$.
		\EndFor
	\end{algorithmic}
\label{alg:training algorithm}
\end{algorithm}

%% file: supp/3_additional_quantitative.tex
\section{Additional Quantitative Results}
\subsection{More Results on NR3D}
\input{supp/figures/supp_quantitative_nr3d}
\input{tables/nr3d_detailed_10percent_performance}

In Sec.~\textcolor{red}{4}, we have shown comparisons between Vigor and several state-of-the-arts under data efficient scenarios (1\% $\sim$ 10\% of data). Here, we provide a comparison between Vigor and MVT+CoT3DRef~\cite{bakr2023cot3dref} from 1\% to 100\% of NR3D data in Fig.~\ref{fig:supp nr3d quant}, where Vigor outperforms CoT3DRef with a noticeable margin when the amount of data is limited and is comparable to CoT3DRef when the data amount is over 30\%, showing that our Vigor is suitable under different settings.

Table~\ref{tab:nr3d detailed 10-percent} further displays the detailed performance on different official subsets of NR3D. Among the subsets, the \textbf{Hard} subset contains samples with more than 2 distractors, where a distractor is an object having the same class name as the target object, and the \textbf{Easy} subset is the contrary. \textbf{View-dependent} samples contain relations where rotating the point cloud scene will affect the referred ground-truth target object (e.g., left and right), and \textbf{View-independent} samples are contrary. Vigor accentuates itself with decent capabilities on different subsets, consistently followed by the two variants of CoT3DRef~\cite{bakr2023cot3dref} that also feature the concept of referential order for 3D visual grounding.

\subsection{More Results on SR3D}
\label{sup: sr3d results}
\input{tables/sr3d_main_table}

Table~\ref{tab:sr3d main table} shows the quantitative comparisons on the SR3D dataset against BUTD-DETR, MVT, CoT3DRef and NS3D~\cite{hsu2023ns3d}, with the settings of using  $1 \%$ and $100\%$ of training examples, respectively. From this table, we can see that although NS3D achieves the best performance when using only $1\%$ of training data due to its structured decomposition of the input description into nested logical expressions, its performance saturates when using more data for training. As for CoT3DRef, its design of applying rule-based matching of anchor/target objects as additional guidance for the model is effective on SR3D when using a large amount of data, where the relations are much simpler than NR3D or ScanRefer, but its performance on $1\%$ of data is suboptimal. On the contrary, our Vigor achieves comparable results in both settings, showing that our design is suitable for various amounts of training pairs.

%% file: supp/figures/supp_quantitative_nr3d.tex
\begin{figure}[tb]
  \centering
  \includegraphics[width=0.45\textwidth]{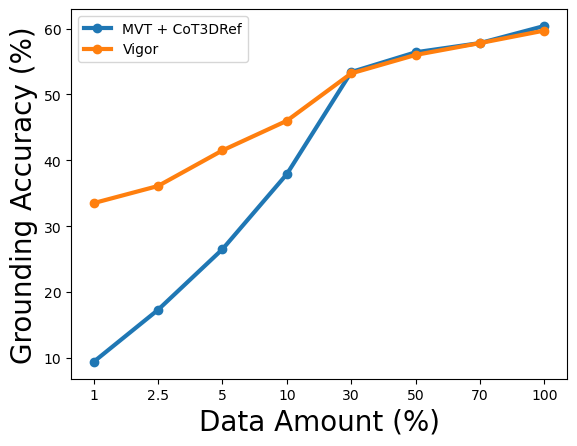}
  \caption{\textbf{Quantitative results on NR3D} We can see that when the amount of data is relatively many (above 30\%), our Vigor is comparable to MVT+CoT3DRef~\cite{bakr2023cot3dref}. However, as the amount of data reduces, our Vigor performs better over MVT-CoT3DRef.}
  \label{fig:supp nr3d quant}
\end{figure}

%% file: tables/nr3d_detailed_10percent_performance.tex
\begin{table*}[tb]
  \caption{\textbf{Grounding accuracy on the official NR3D subsets~\cite{achlioptas2020referit3d}.} For implementation and comparison purposes, only the setting of 10\% of training data is considered. 
  }
  \label{tab:nr3d detailed 10-percent}
  \centering
  \addtolength{\tabcolsep}{8pt}
  \resizebox{0.75\linewidth}{!}{
  \begin{tabular}{@{}l|ccccc@{}}
    \toprule
     Method & Hard & Easy & View-Dep. & View-Indep. & Overall \\
    \midrule
    Referit3D~\cite{achlioptas2020referit3d} & 19.5 & 27.3 & 21.2 & 24.2 & 23.3 \\
    TransRefer3D~\cite{he2021transrefer3d} & 21.6 & 29.9 & 22.9 & 27.0 & 25.7 \\
    SAT~\cite{yang2021sat} & 22.5 & 27.6 & 21.7 & 26.6 & 25.0 \\
    BUTD-DETR~\cite{jain2022bottom} & {25.9} & {41.9} & {29.1} & {34.8} & {33.3} \\
    MVT~\cite{huang2022multi} & 22.9 & 30.3 & 25.4 & 27.1 & 26.5 \\
    MVT + CoT3DRef~\cite{bakr2023cot3dref} &  \underline{32.7} & 43.2 & \underline{34.0} & 39.8 & 37.9 \\
    ViL3DRel + CoT3DRef~\cite{bakr2023cot3dref} &  32.4 & \underline{44.7} & 33.4 & \underline{40.9} & \underline{38.4} \\
    Vigor (Ours) & \textbf{39.1} & \textbf{53.3} & \textbf{45.3} & \textbf{46.4} & \textbf{46.0} \\
  \bottomrule
  \end{tabular}}
\end{table*}

%% file: tables/sr3d_main_table.tex
\begin{table}[tb]
  \caption{\textbf{Data Efficient Grounding accuracy (\%) on SR3D.} We show the results trained with 1\% and 100\% of training data.}
  \vspace{-0mm}
  \label{tab:sr3d main table}
  \centering
  \addtolength{\tabcolsep}{8pt}
  \resizebox{1\linewidth}{!}{
  \begin{tabular}{@{}l|ccc@{}}
    \toprule
    \multirow{2}{*}{Method} & \multicolumn{3}{c}{Labeled Training Data} \\
    \cline{2-4}
    &1\%  & 100\% \\
    \midrule
    BUTD-DETR~\cite{jain2022bottom} & 36.5  & 67.0 \\
    MVT~\cite{huang2022multi} & 22.2  & 64.5\\
    NS3D~\cite{hsu2023ns3d} & \textbf{52.4}  & 62.7 \\
    MVT + CoT3DRef~\cite{bakr2023cot3dref} & 26.9  & \textbf{73.2}\\
    Vigor (Ours) & \underline{51.3}  & \underline{67.1}\\
  \bottomrule
  \end{tabular}}
\end{table}


%% file: supp/4_loss_ablation.tex
\section{Ablation Studies on Training Objectives}
\label{sup:loss ablation}
This section performs the ablation study on several training objectives used for Vigor mentioned in Sec.~\textcolor{red}{3}. In particular, we investigate the influence of $\mathcal{L}_{mask}$, $\mathcal{L}_{crd}$, and $\mathcal{L}_{text}$ for low-resource (1\% training data) and full-resource (100\% training data) scenarios, as shown in Table~\ref{tab:proposed losses ablations}. Note that we only conduct target object classification on the text feature $T$ when deactivating $\mathcal{L}_{text}$, and apply classification on both anchor and target objects when adopting $\mathcal{L}_{text}$, as mentioned in Sec.~\textcolor{red}{3.3}. Also, for all experiments in Table~\ref{tab:proposed losses ablations}, the pre-training with synthetic data in Sec.~\textcolor{red}{3.4} is applied for fair comparison. In summary, $\mathcal{L}_{mask}$, $\mathcal{L}_{crd}$, and $\mathcal{L}_{text}$ all impose positive effects on models' performances under the two settings, verifying Vigor's design in Sec.~\textcolor{red}{3}.

\begin{table*}[tb]
  \caption{\textbf{Ablation studies on training objectives.} Note that for all ablation settings, the proposed order-aware pre-training in Sec.~\textcolor{red}{3.4} is applied for fair comparison.}
  \label{tab:proposed losses ablations}
  \centering
  \addtolength{\tabcolsep}{8pt}
  \begin{tabular}{@{}cccc|cc@{}}
    \toprule
    $\mathcal{L}_{mask}$ & $\mathcal{L}_{crd}$ & $\mathcal{L}_{text}$ & $\mathcal{L}_{ref}$ & 1$\%$ & 100$\%$ \\
    \midrule
     &  &  & \ding{51} & 31.1 & 54.2 \\
     & & \ding{51} & \ding{51} & 31.6 & 54.8 \\
     & \ding{51} & \ding{51} & \ding{51} & 32.4 & 56.0 \\
    \ding{51} & \ding{51} & \ding{51} & \ding{51} & 33.5 & 59.7 \\
  \bottomrule
  \end{tabular}
\end{table*}

%% file: supp/6_order_length_as_subsets.tex
\section{Model Performance Under Different Order Length}
\label{sup:order length as subsets}
\input{supp/figures/order_length_ablation}
\begin{table*}[tb]
  \caption{\textbf{Grounding accuracy on NR3D subsets regarding different parsed referential order lengths.} Note that referential order length $=1$ means only the target object is mentioned in the description. The improvement of Vigor over MVT~\cite{huang2022multi} grows as the parsed order length increases.}
  \label{tab:order length as subsets}
  \centering
  \addtolength{\tabcolsep}{8pt}
  \begin{tabular}{@{}c|ccc|c@{}}
    \toprule
    \multirow{2}{*}{Method} & \multicolumn{3}{c|}{Order Length}  & \multirow{2}{*}{overall} \\
    \cline{2-4}
    & 1& 2\&3 & 4\&5 & \\
    \midrule
     MVT~\cite{huang2022multi} & 59.4 & 55.0 & 46.7 & 55.1 \\
     \midrule
     Vigor (Ours)& \makecell[c]{61.6\\(+2.2)} & \makecell[c]{59.6\\(+4.6)} & \makecell[c]{52.0\\(+5.3)} & \makecell[c]{59.7\\(+4.6)} \\
  \bottomrule
  \end{tabular}
\end{table*}
In this section, we analyze the effectiveness of padded order length $B$ of $O_{1:B}$ (and also the number of Referring Blocks) in Fig.~\ref{fig:order length ablation}, which illustrates the performances with different $B$ using 1\% and 100\% training data in Fig.~\ref{fig:order length performance}. We also show the statistics of the original order length generated by our two-stage referential ordering in Fig.~\ref{fig:order length statistics}. We observe that although performances gradually increase with larger $B$ (i.e., considering more potential anchor objects appeared in $D$ achieves better accuracy), the performance gain saturates at $B = 4$. This can be explained by looking at Fig.~\ref{fig:order length statistics}, which shows that only a very small amount of data exceeds an order length of $4$ in both training and testing pairs. This suggests that our selection of $B = 4$ is reasonable, optimally balancing computational efficiency with prediction accuracy.

We additionally explore Vigor's performance gain for descriptions with different order lengths compared with MVT~\cite{huang2022multi} to show the effectiveness of our design of progressive location to the target object. We split the original NR3D testing set into subsets that possess order lengths of \textbf{1}, \textbf{2\&3}, and \textbf{4\&5}. An order length of 1 (1092 samples in the testing set) means that only the target object is mentioned in the description. Samples with an order length of 2 or 3 (6184 samples in the

\noindent testing set) have 1 or 2 anchor objects mentioned other than the target object. Similarly, samples with an order length of 4 or 5 (209 samples in the testing set) have 3 or 4 anchor objects mentioned other than the target object.
A longer referential order may generally denote a longer and more complicated description. As shown in Table~\ref{tab:order length as subsets}, though only a 2.2\% performance gain is obtained for target-only samples, Vigor is more advantageous when dealing with lengthy descriptions, with 5.3\% performance gain achieved for descriptions with order length of 4 or 5.

%% file: supp/figures/order_length_ablation.tex
\begin{figure*}[tb]
  \centering
  \begin{subfigure}{0.45\linewidth}
    \includegraphics[width=\textwidth]{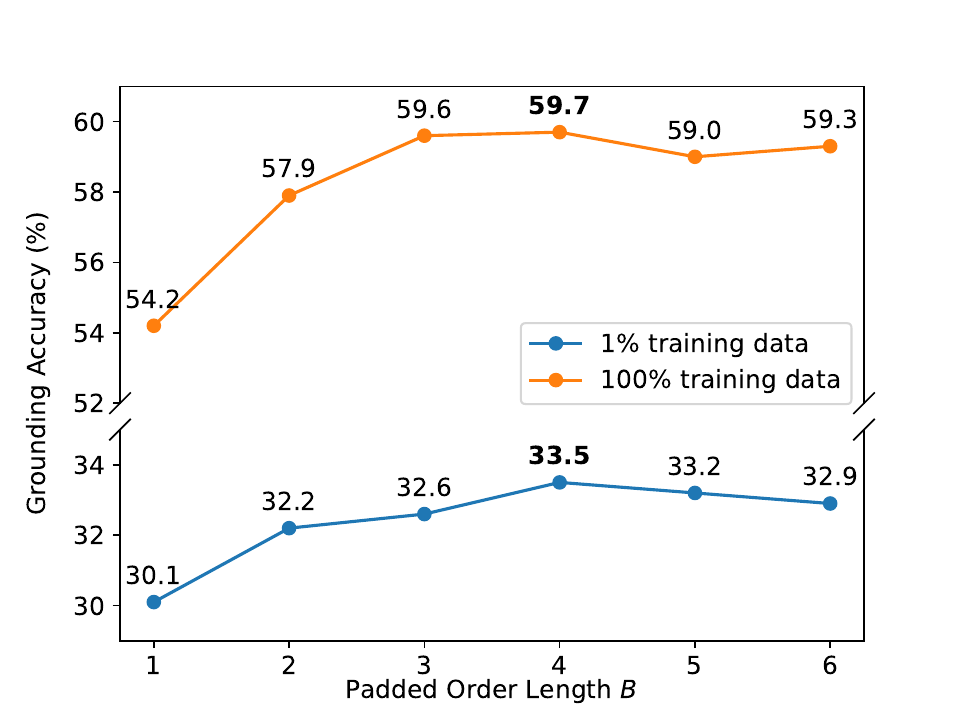}
    \caption{\textbf{Grounding accuracy with different maximum (padded) referential order length $B$.} Note that $B$ is equal to and denoted as the number of Object-Referring blocks.}
    \label{fig:order length performance}
  \end{subfigure}
  \hfill
  \begin{subfigure}{0.45\linewidth}
    \includegraphics[width=\textwidth]{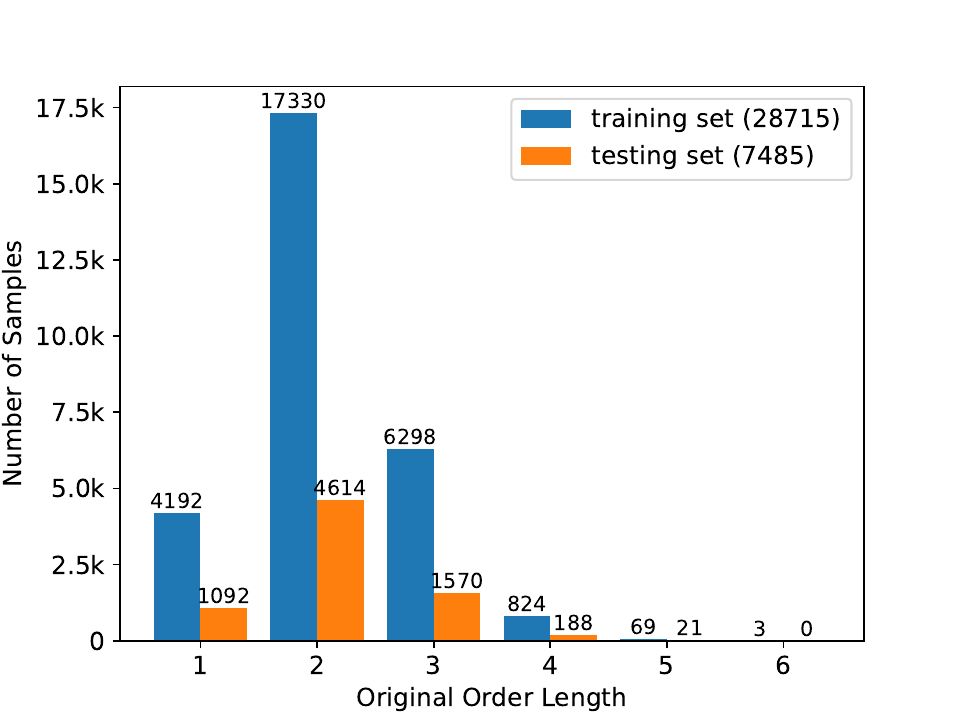}
    \caption{\textbf{Statistics of the referential order length in NR3D.} A longer referential order denotes a more complex referring process for grounding.}
    \label{fig:order length statistics}
  \end{subfigure}
  \caption{\textbf{(a) Grounding accuracies with varying Object-Referring block numbers $B$ and (b) statistics of referential order length of NR3D.} It can be seen that the model performance saturates at $B$=4, matching length statistics of NR3D.}
  \label{fig:order length ablation}
\end{figure*}

%% file: supp/5_more_visualization.tex
\section{Visualization of Responses in Each Referring Block}
\label{sup:more visualization}
\input{supp/figures/supp_qualitative_figure}
To show that our Vigor indeed progressively locates the target object following the derived referential order, we visualize the feature response of $F_{1:(B+1)}$ ($B=4$) in Fig.~\ref{fig:supp qualitative results}. The blue bounding box indicates the ground-truth target object, and we color the object proposal according to the response of their corresponding features in $F_{1:(B+1)}$, where a brighter color represents a higher response. We can see that the responses to object proposals are originally cluttered. As our referential blocks are applied, the response of anchor/target objects becomes larger and finally locates the ideal target object in the last feature $F_5$.

%% file: supp/figures/supp_qualitative_figure.tex
\begin{figure*}[tb]
  \centering
  \includegraphics[width=0.8\textwidth]{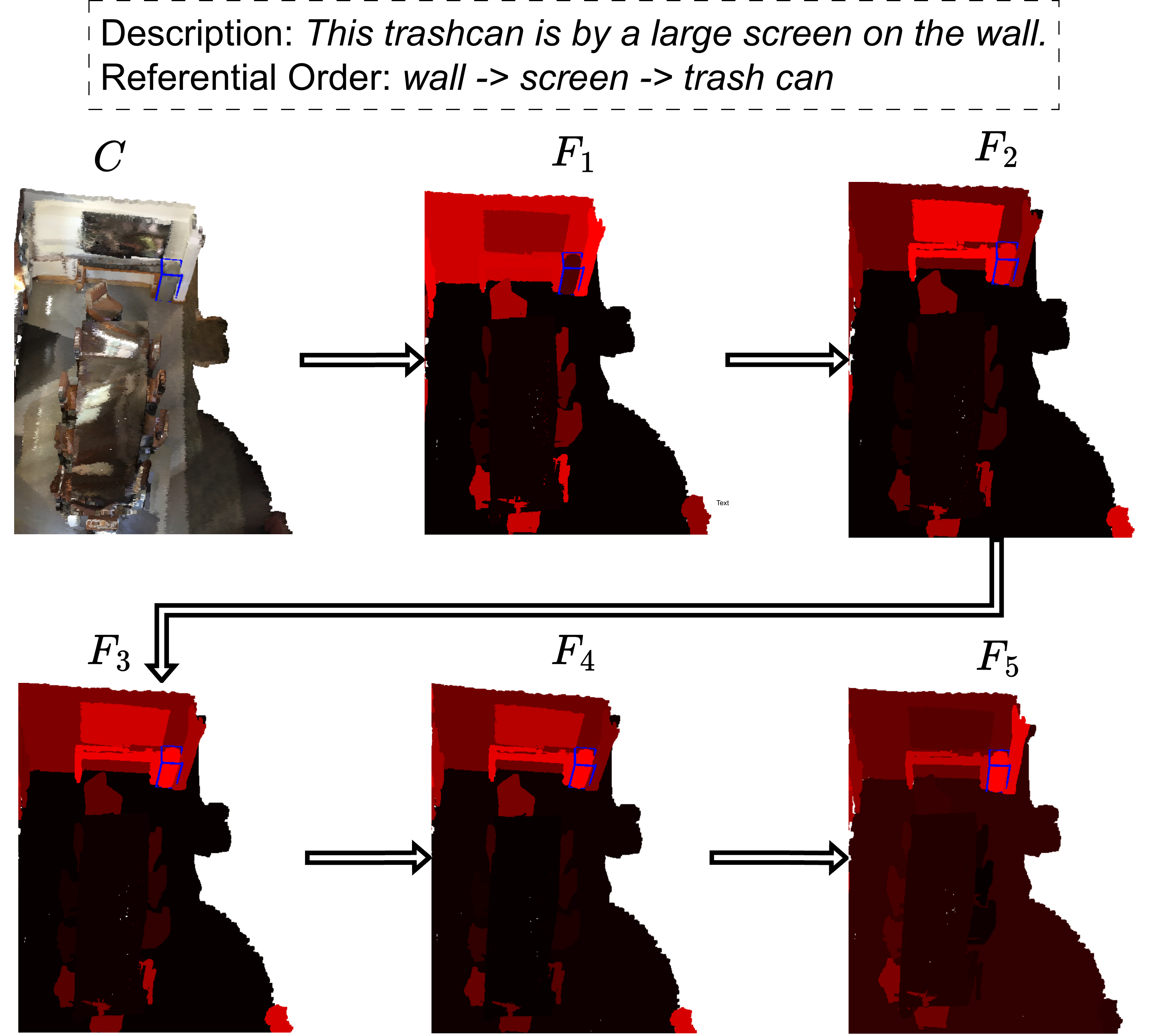}
  \caption{\textbf{Visualization of Responses in Each Referring Block.} An example in NR3D is shown in this figure. We color each object proposal according to their feature response in $F_{1:(B+1)}$, where a larger response with a brighter color. Note that the blue bounding box represents the ground truth target object. We can see that our Vigor progressively locates the target object by considering the referential order by first focusing on the wall then the screen and finally the trash can.}
  \label{fig:supp qualitative results}
\end{figure*}

%% file: supp/2_prompt_and_examples.tex
\section{Details and Prefix Prompt Examples of Two-Stage ICL for Deriving Referential Order}
\label{sup:prompts}
\input{supp/figures/supp_llm}
To have Object-Referring blocks $\{R_1, \cdots, R_B\}$ in Sec.~\textcolor{red}{3} to locate the target object properly, it is desirable to extract a proper referential order $O_{1:B}$ from the input description $D$. With such a referring path constructed, visual features of the associated objects can be updated for grounding purposes. This is inspired by the idea of Chain-of-Thoughts~\cite{wei2022chain, kojima2022large} in LLM, as noted in CoT3DRef~\cite{bakr2023cot3dref}. To achieve such a parsing task, we apply GPT-3.5-Turbo~\cite{gpt35turbo} as the description parser using in-context learning (ICL)~\cite{dong2022incontextlearning}, as depicted in Fig.~\ref{fig:supp llm}.

However, it is not trivial to have LLM output a referential order from $D$ due to lengthy and noisy descriptions. For example, for an input description ``\textit{Look at the king-size bed in the room next to a green chair, Find the pillow on the bed. Not the pillow on the sofa beside the chair.}'', one would expect first to find the green \textit{chair} then the king-size \textit{bed} next to the chair, and finally, the \textit{pillow} on the bed. Therefore, the ideal referential order is \textit{\{``chair'', ``bed'', ``pillow''\}}. In the above example, the sentence ``\textit{Not the pillow on the sofa which is also beside the chair.}'' is redundant since one can find the correct target object without this information. If we apply the LLM directly to the original description, the model may be misled by this redundant information and generate a referential order containing the object ``\textit{sofa}''.

To tackle the above problem, we conduct a two-stage in-context learning (ICL) scheme to remove redundant information in $D$ before producing $O_{1:B}$, as depicted in Fig.~\ref{fig:supp llm}.

With the given target object $O_B$, we predict a summarized description $D^{'}$ to remove redundant information in $D$ in the first stage. Then, the entire referential order $O_{1:B}$ given $D^{'}$ and $O_B$ is produced in the second stage.
For each stage of our ICL, \textit{10} examples are provided as demonstrations of the input prompts for LLM to predict $O_{1:B}$. Due to page limitations, demonstration examples and the complete prefix prompts are presented in the supplementary materials. With $O_{1:B}$ obtained, the order-aware object referring process can be processed accordingly, as we detail next.

We now list the complete prompts of our two-stage ICL using GPT-3.5-Turbo~\cite{gpt35turbo} for referential order generation. Also, we exhibit parsing results of 4 samples in the NR3D testing set and compare them with CoT3DRef~\cite{bakr2023cot3dref} that also establishes the extraction and usage of referential order.

\subsection{Prefix Prompt of First-Stage ICL}
The first-stage prompt is used to acquire the summarized description $D^{'}$ and target object $O_B$ of the original description. Our first-stage prompt is as follows:

\textit{I have some descriptions, each describing a specific target object in a room. However, they may have some redundant clauses or words. Your task is to summarize them into a shorter description. Also, tell me what the target object.\\Below are 10 examples:\\}
\noindent
\textit{\textbf{description 1}: Assume you are facing the door in the room. Find the larger cabinet to its left.} \\
\textit{\textbf{summarized description 1}: When facing the door, the cabinet on the right of it.}\\
\textit{\textbf{target object 1}: cabinet}
\\
\\
\textit{\textbf{description 2}: The water bottle that is above the easy chair. NOT the smaller water bottle that is above the orange table.} \\
\textit{\textbf{summarized description 2}: The water bottle that is above the easy chair.}\\
\textit{\textbf{target object 2}: water bottle}
\\
\\
\textit{\textbf{description 3}: In the bedroom, you will see a sheer curtain. Beside the curtain is the steel window you need to find.} \\
\textit{\textbf{summarized description 3}: The steel window beside a sheer curtain.}\\
\textit{\textbf{target object 3}: window}
\\
\\
\textit{\textbf{description 4}: Please find the towel hanging on the wall in the bathroom with the other three towels. You should find the one nearest to the door. Or say it is on the door’s right side.} \\
\textit{\textbf{summarized description 4}: The towel on the wall nearest to the door.}\\
\textit{\textbf{target object 4}: towel}
\\
\\
\textit{\textbf{description 5}: Between a pencil and a desk lamp on the desk is the backpack you need to find.} \\
\textit{\textbf{summarized description 5}: The backpack between a pencil and a desk lamp on the desk.}\\
\textit{\textbf{target object 5}: backpack}
\\
\\
\textit{\textbf{description 6}: In the living room we have three bookshelves. Choose the bookshelf to the right of the clock facing a cabinet.} \\
\textit{\textbf{summarized description 6}: The bookshelf to the right of the clock faces a cabinet.}\\
\textit{\textbf{target object 6}: bookshelf}
\\
\\
\textit{\textbf{description 7}: The person wearing a white T-shirt, not the man who is also sitting on the bed but with a jacket.} \\
\textit{\textbf{summarized description 7}: The person wearing a white T-shirt on a bed.}\\
\textit{\textbf{target object 7}: person}
\\
\\
\textit{\textbf{description 8}: The purple pillow on the right side of the bed when facing it. Not the one on the left side and the one in the middle of the bed.} \\
\textit{\textbf{summarized description 8}: The purple pillow on the right side of the bed when facing it.}\\
\textit{\textbf{target object 8}: pillow}
\\
\\
\textit{\textbf{description 9}: The brown door at the end of the living room, next to the trash cans, which are full of garbage.} \\
\textit{\textbf{summarized description 9}: The brown door next to the full trash can.}\\
\textit{\textbf{target object 9}: door}
\\
\\
\textit{\textbf{description 10}: The shoes that are placed in the middle of five shoes near the door in the room.} \\
\textit{\textbf{summarized description 10}: The middle shoes near the door.}\\
\textit{\textbf{target object 10}: shoes}
\\
\textit{Now for the description [DESCRIPTION], give me the summarized description and the target object. Your answer must be in the form "summarized description:
target object:"}\\
\subsection{Prefix Prompt of Second-Stage ICL}
The second-stage prompt is used to acquire the referential order $O_{1:B}$ based on the target object $O_B$ and the summarized description $D'$. The complete prompt is as follows:

\textit{I have some descriptions, each describing a specific target object with some supporting anchor objects helping the localization. We can find the specific target object by tracing the referential order of anchor objects step by step. Your task is to provide a correct referential order. Also, tell me what the mentioned anchor objects.\\Below are 10 examples:\\}
\noindent
\textit{\textbf{description 1}: The water bottle that is above the easy chair.} \\
\textit{\textbf{target object 1}: water bottle}\\
\textit{\textbf{anchor objects 1}: easy chair}\\
\textit{\textbf{referential order 1}: easy chair$\rightarrow$water bottle}
\\
\\
\textit{\textbf{description 2}: The steel window beside a sheer curtain.} \\
\textit{\textbf{target object 2}: window}\\
\textit{\textbf{anchor objects 2}: curtain}\\
\textit{\textbf{referential order 2}: curtain$\rightarrow$window}
\\
\\
\textit{\textbf{description 3}: The trash can that is on the right of the king-size bed.} \\
\textit{\textbf{target object 3}: trash can}\\
\textit{\textbf{anchor objects 3}: bed}\\
\textit{\textbf{referential order 3}: bed$\rightarrow$trash can}
\\
\\
\textit{\textbf{description 4}: The backpack between a pencil and a desk lamp. They are all on a wooden desk.} \\
\textit{\textbf{target object 4}: backpack}\\
\textit{\textbf{anchor objects 4}: pencil, desk lamp, desk}\\
\textit{\textbf{referential order 4}: desk$\rightarrow$pencil$\rightarrow$desk lamp$\rightarrow$backpack}
\\
\\
\textit{\textbf{description 5}: The cabinet on the right of the door.} \\
\textit{\textbf{target object 5}: cabinet}\\
\textit{\textbf{anchor objects 5}: door}\\
\textit{\textbf{referential order 5}: door$\rightarrow$cabinet}
\\
\\
\textit{\textbf{description 6}: The bookshelf to the right of the clock facing a cabinet.} \\
\textit{\textbf{target object 6}: bookshelf}\\
\textit{\textbf{anchor objects 6}: clock, cabinet}\\
\textit{\textbf{referential order 6}: cabinet$\rightarrow$clock$\rightarrow$bookshelf}
\begin{table*}[tb]
  \caption{\textbf{Examples of LLM-parsed referential order from CoT3DRef~\cite{bakr2023cot3dref} and Vigor.} Note that the blue text represents the ideal anchor/target objects, and the red text represents the redundant object that should not appear in the referential order. We can see that MVT misses one anchor object in the third example and includes a redundant object in the fourth example, while Vigor predicts proper order for both cases.}
  \label{tab:parsing examples}
  \centering
  \addtolength{\tabcolsep}{4pt}
  \begin{tabular}{@{}p{0.42\textwidth}|p{0.25\textwidth}|p{0.25\textwidth}@{}}
    \toprule
    Description & CoT3DRef~\cite{bakr2023cot3dref} & Vigor (Ours)\\
    \midrule
    \makecell[l]{The \textcolor{blue}{pillow} closest to the foot of\\ the \textcolor{blue}{bed}.} & \makecell[l]{\textcolor{blue}{bed} $\rightarrow$ \textcolor{blue}{pillow}} & \makecell[l]{\textcolor{blue}{bed} $\rightarrow$ \textcolor{blue}{pillow}}\\
    \midrule
    \makecell[l]{Facing the \textcolor{blue}{bed}, it's the large white\\ \textcolor{blue}{pillow} on the right. The second \\one from the \textcolor{blue}{headboard}.} & \makecell[l]{\textcolor{blue}{bed} $\rightarrow$ \textcolor{blue}{headboard}\\$\rightarrow$ \textcolor{blue}{pillow}} & \makecell[l]{\textcolor{blue}{bed} $\rightarrow$ \textcolor{blue}{headboard}\\$\rightarrow$ \textcolor{blue}{pillow}}\\
    \midrule
    \makecell[l]{The front \textcolor{blue}{pillow} on the \textcolor{blue}{bed} with\\ the \textcolor{blue}{laptop}.} & \makecell[l]{\textcolor{blue}{bed} $\rightarrow$ \textcolor{blue}{pillow}} & \makecell[l]{\textcolor{blue}{laptop} $\rightarrow$ \textcolor{blue}{bed}\\$\rightarrow$ \textcolor{blue}{pillow}}\\
    \midrule
    \makecell[l]{The \textcolor{blue}{window} near the \textcolor{blue}{table}, not\\ the one near the \textcolor{red}{shelves}.} & \makecell[l]{\textcolor{blue}{table} $\rightarrow$ \textcolor{red}{shelves}\\$\rightarrow$ \textcolor{blue}{window}} & \makecell[l]{\textcolor{blue}{table} $\rightarrow$ \textcolor{blue}{window}}\\
    \bottomrule
  \end{tabular}
\end{table*}
\\
\\
\\
\textit{\textbf{description 7}: The person wearing a white T-shirt on a bed.} \\
\textit{\textbf{target object 7}: person}\\
\textit{\textbf{anchor objects 7}: bed}\\
\textit{\textbf{referential order 7}: bed$\rightarrow$person}
\\
\\
\textit{\textbf{description 8}: The purple pillow on the right side of the bed when facing it.} \\
\textit{\textbf{target object 8}: pillow}\\
\textit{\textbf{anchor objects 8}: bed}\\
\textit{\textbf{referential order 8}: bed$\rightarrow$pillow}
\\
\\
\textit{\textbf{description 9}: The brown door next to the full trash can.} \\
\textit{\textbf{target object 9}: door}\\
\textit{\textbf{anchor objects 9}: trash can}\\
\textit{\textbf{referential order 9}: trash can$\rightarrow$door}
\\
\\
\textit{\textbf{description 10}: Please find the towel hanging on the wall in the bathroom with the other three towels. You should find the one nearest to the door. Or say it is on the door’s right side.} \\
\textit{\textbf{target object 10}: towel}\\
\textit{\textbf{anchor objects 10}: wall, door}\\
\textit{\textbf{referential order 10}: wall$\rightarrow$door$\rightarrow$towel}\\

\textit{Now for the description: [DESCRIPTION], give me the anchor objects and the referential order. Your answer must be in the form "referential order, anchor objects:. "}

\subsection{Examples of Derived Referential Order}
To show that our two-stage ICL produces reasonable referential orders, we provide examples and comparisons between ours and CoT3DRef~\cite{bakr2023cot3dref}'s parsing results. In particular, we leverage CoT3DRef's released prompt to query the GPT-3.5-Turbo. We display \textit{4} examples in Table~\ref{tab:parsing examples}, where CoT3DRef misses an anchor object \textit{``bed''} in the third example and includes a redundant object \textit{``shelves''} as an anchor object in the fourth example, while our Vigor produces proper results in both cases. This shows the effectiveness of the two-stage ICL strategy.

%% file: supp/figures/supp_llm.tex
\begin{figure}[tb]
  \centering
  \includegraphics[width=0.45\textwidth]{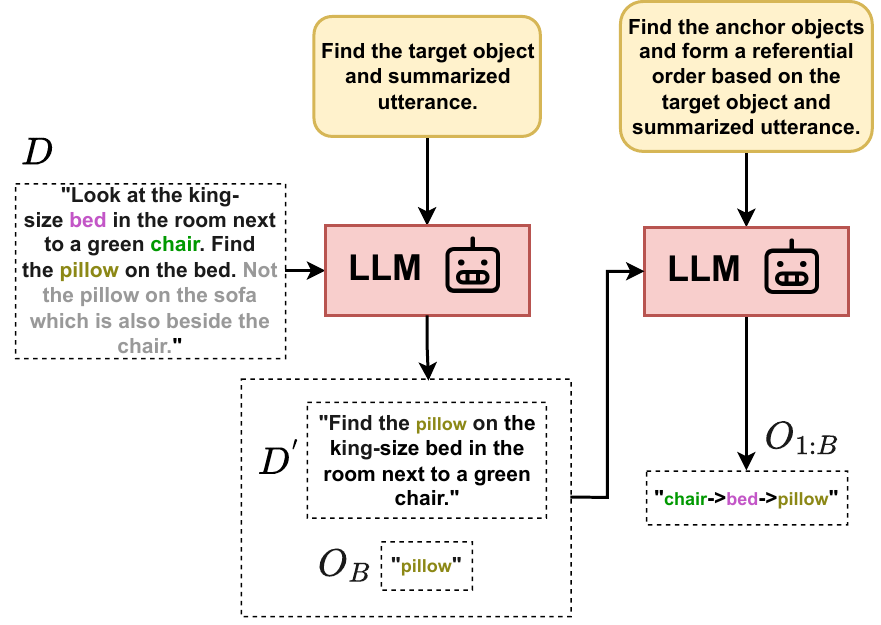}
  \caption{\textbf{Referential order generation via LLM.} A two-stage ICL is deployed to remove redundant information from $D$, forming a descriptive order $O_{1:B}$ for locating the target object $O_B$.}
  \label{fig:supp llm}
\end{figure}

%% file: supp/7_limitations.tex
\section{Limitations and Social Impact}
\label{supp:limitation}
\subsection{Limitations}
\input{tables/parsing_stat}

\subsubsection{LLM-parsed Referential Order}
Since Vigor utilizes LLMs to generate the referential order from the input description, despite of our introduced pre-training strategy to warm the training process, the correctness of the extracted referential order would affect the training and the performance of Vigor. For example, Table~\ref{tab:parsing stat} shows the accuracy on the SR3D dataset and NR3D dataset that our GPT-3.5-Turbo-parsed referential orders correctly place the target object at the last position. We can see that for the NR3D dataset, where the relations in the descriptions are much more complicated, the accuracy of the identification of the target objects is $86.9\%$ and $89.1\%$ for the training and testing sets, respectively. The gap between this accuracy and an absolutely reliable prediction (i.e., $100\%$ accuracy), though small, would still affect the training stability and testing accuracy of our visual grounding pipeline. As a result, better usage of the LLM to produce perfect referential order is one of the future research directions to pursue.
\subsubsection{Order Length Decision}
As detailed in Sec.~\ref{sup:implementation details}, we set the length $B$ of referential order (as well as the number of referring blocks) as $4$ to conduct batch-wise training. Although we have shown that our choice of $B$ is reasonable for balancing computational efficiency and prediction accuracy in Sec.~\ref{sup:order length as subsets} by conducting proper experiments on NR3D, this choice appears to be dataset-specific. We leave this as a future direction to develop a more flexible architecture to be able to dynamically adjust $B$ according to the LLM-parsed referential order for each sample during training and testing.

\subsection{Broader Impact}
\label{broader impact}
\subsubsection{Applications of 3D visual grounding}
Although the experiments in this paper are conducted on indoor datasets only, the task of 3D visual grounding is not restricted to indoor scenes. For the autonomous driving industry, 3D visual grounding could also be an important topic to study along with 3D object detection. As our approach is designed to promote the data efficiency of 3D visual grounding tasks, we look forward to seeing future works that consider the concept of our Vigor and apply it to autonomous driving systems since it is hard to collect enormous amounts of data for autonomous driving.

\subsubsection{Potential Negative Impacts}
Although the experiments in this paper show that our Vigor outperforms current SOTAs in data-efficient scenarios, one must make sure that Vigor is well-validated before applying it to a new data domain. Without properly transferring Vigor's grounding ability to the corresponding domain, the performance could be non-ideal, introducing potential safety risks, especially in critical applications like autonomous driving.

%% file: tables/parsing_stat.tex
\begin{table*}[t]
  \caption{\textbf{Our adopted GPT-3.5-Turbo's zero-shot accuracy (\%) on identifying the class name of the target object in the referring descriptions.} Note that since the ground-truth labels and orders of the anchor objects are not available, we are only allowed to check if the target object is correctly placed at the last position in the parsed referential order as an indirect verification of the reliability of the orders.}
  \vspace{-0mm}
  \label{tab:parsing stat}
  \centering
  \addtolength{\tabcolsep}{8pt}
  \resizebox{0.5\linewidth}{!}{
  \begin{tabular}{@{}l|cc|cc}
    \toprule
    \multirow{2}{*}{Method} & \multicolumn{2}{c}{NR3D} & \multicolumn{2}{c}{SR3D} \\
    \cline{2-5}
    & train & test & train & test \\
    \midrule
    GPT-3.5-Turbo & 86.9 & 89.1 & 96.4 & 96.1\\
  \bottomrule
  \end{tabular}}
\end{table*}